\documentclass{article}

\PassOptionsToPackage{numbers, compress}{natbib}
\usepackage[preprint]{neurips_2026}

\usepackage[utf8]{inputenc} % allow utf-8 input
\usepackage[T1]{fontenc}    % use 8-bit T1 fonts
\usepackage{hyperref}
\hypersetup{
    hidelinks,
    colorlinks=true,
}
\usepackage{url}            % simple URL typesetting
\usepackage{booktabs}       % professional-quality tables
\usepackage{amsfonts}       % blackboard math symbols
\usepackage{nicefrac}       % compact symbols for 1/2, etc.
\usepackage{microtype}      % microtypography
\usepackage{xcolor}         % colors
\usepackage{amsthm}
\usepackage{amsmath}
\usepackage{xspace}
\usepackage{graphicx}
\usepackage{algorithm}
\usepackage{algpseudocode}

\theoremstyle{plain}
\newtheorem{theorem}{Theorem}

\newtheorem{proposition}[theorem]{Proposition}

\theoremstyle{definition}
\newtheorem{definition}[theorem]{Definition}

\newcommand\sys{\texttt{DEX}\xspace}% System name to be decided

\algrenewcommand{\algorithmiccomment}[1]{\hfill\textcolor{blue}{\(\triangleright\) #1}}
\newcommand{\fn}[1]{\mathsf{#1}}
\title{Depth Exploration for LLM Decoding}

\author{%
  Weisi Yang \\
  Northwestern University \\
  \And
  Zipeng Sun \\
  MILA -- Quebec AI Institute \\
  McGill University \\
  \And
  Stephen Xia \\
  Northwestern University \\
}

\begin{document}

\maketitle

\begin{abstract}
Autoregressive LLM decoding evaluates every generated token through the full layer stack, even though many tokens become predictable at intermediate depths. Existing lossless depth-adaptive methods exploit this redundancy by choosing a single non-final exit depth and verifying its prediction with the final-depth model. However, our measurements show that this selection-based strategy leaves substantial headroom: choosing an exit too late wastes computation, while choosing one too early triggers fallback and discards dependent drafts. We propose Depth Exploration Decoding (DEX), a lossless decoding algorithm that replaces single-depth selection with parallel exploration over multiple candidate depths. At each commit position, DEX validates candidates against the final-depth reference, commits exactly the final-depth token, and collapses the exploration lattice to retain only reusable branch states. This expand--commit--collapse procedure preserves equivalence to standard autoregressive decoding while reducing the cost of committing each token. Across early-exit-trained and standard LLMs, DEX outperforms representative depth-selection baselines and achieves competitive end-to-end throughput against speculative and distributed decoding methods. Moreover, DEX improves as the explored depths become finer, showing that parallel depth exploration provides a scalable way to exploit the underused depth axis of LLM decoding. 
\end{abstract}

\section{Introduction}

The large parameter size of modern Large Language Models (LLMs) gives them strong capabilities across many domains, but also makes decoding expensive. 
A major source of this latency is autoregressive (AR) decoding: LLMs generate responses token by token, and each generated token must pass through the full layer stack before the next token can be produced. 
Recent studies~\cite{lioubashevski2025looking,csordas2026do} identify redundancy along the depth dimension, suggesting that not all tokens require the full model depth to become predictable. 
Leveraging this observation, prior work~\cite{Elbayad2020Depth-Adaptive,schuster2022confident,chen2024eellm} has proposed depth-adaptive decoding methods that allow tokens to exit from intermediate layers, reducing the amount of layer computation and improving decoding latency.

More recently, the draft-and-verify paradigm in speculative decoding~\cite{leviathan2023fast} has been adopted by lossless depth-adaptive methods~\cite{elhoushi2024layerskip, wei2025adadecode,zarch2025del}. 
These methods draft candidates from non-final depths and verify them with the final-depth model, preserving equivalence with standard AR decoding while reducing layer computation. 
However, existing methods share a \emph{selection bottleneck} that limits their potential speedup: they choose only a single exit depth for each token, either through a fixed or learned policy. 

Choosing too late wastes layers even when the token was already predictable earlier, while choosing too early can trigger final-depth replacement and discard descendant drafts. 
Our empirical studies confirm this gap between existing methods and an oracle that leverages the earliest intermediate depth whose prediction matches the full-depth output (Figure~\ref{fig:depth_speed}). 
This raises the key question: \emph{can we exploit depth redundancy by reducing the risk of a wrong exit-depth decision?}

We propose depth exploration (\sys), whose key idea is to replace single-depth prediction with multi-depth coverage. 
Instead of betting on one exit depth, \sys explores several candidate depths in parallel and uses the full-depth output as the reference to identify the earliest explored branch that matches it. 
A missed shallow candidate therefore does not immediately force a full-depth fallback: a deeper explored branch may still match and be reused. 
As a result, the error of depth adaptation is controlled by the spacing between explored depths rather than by the accuracy of a single exit-depth predictor. 
As the exploration set becomes finer, the first reusable branch moves closer to the earliest depth at which the token would have matched the full-depth output.

The main challenge is not merely exposing more exits, but turning parallel depth coverage into an exact decoding procedure. 
Exploration creates multiple possible futures, yet AR decoding has only one official prefix. 
After each commit, the decoder must identify which branch computations remain valid under the new prefix and which have become stale. 
\sys addresses this with an expand--commit--collapse procedure: it \textbf{expands} candidate branches across depth stages, \textbf{commits} only the final-depth-validated token, and \textbf{collapses} the exploration lattice onto the branch consistent with that commit (Section~\ref{sec:lattice_fsm}). 
This allows \sys to reuse parallel depth exploration when it agrees with the final model, while preserving lossless equivalence to standard AR decoding.

To conclude, our main contributions are:

\begin{itemize}
    \item We define Earliest Available Depth (EAD) to quantify depth-side token readiness. Building on EAD, we formulate depth selection and depth exploration, and identify the selection bottleneck in existing lossless depth-adaptive decoding methods.
    \item We present \sys, a lossless decoding algorithm that instantiates depth exploration through parallel branch expansion, final-depth validation and commit, and collapse for clearing tokens from rejected branches.
    \item We show that \sys outperforms prior depth-based baselines and achieves competitive end-to-end throughput against representative speculative and distributed decoding baselines, with stronger scaling as more depth explorers are available.
\end{itemize}

\begin{figure}[t]
    \centering
    \includegraphics[width=\linewidth]{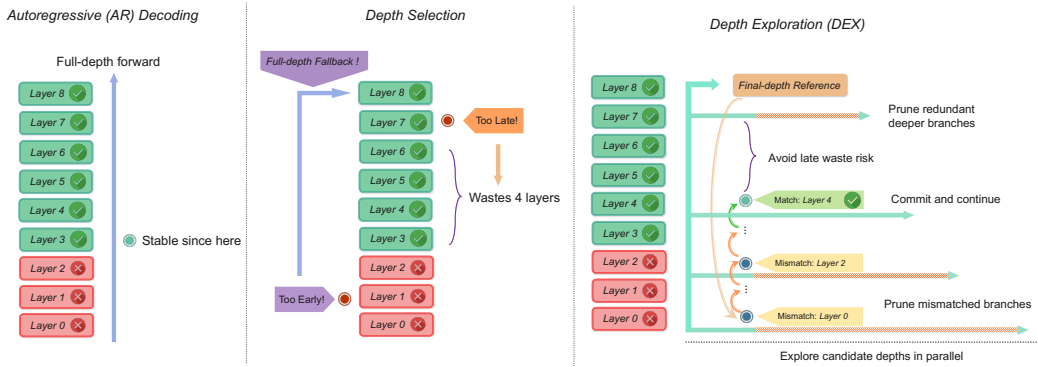}

    \caption{
    Comparison of different depth exploitation.
    (Left) Autoregressive decoding always performs a full-depth forward pass.
    (Middle) Depth selection relies on a single exit-depth decision, risking either full-depth fallback when exiting too early or wasted computation when exiting too late.
    (Right) Parallel depth exploration over candidate depths resolves branches with the final-depth reference and preserves only reusable branches.
        }
    \label{fig:mechanism}
\end{figure}
\section{Preliminaries and problem formulation}
\label{sec:prelim}

This section gives the compact formulation used by the method and experiments. Appendix~\ref{app:formulation} provides additional EAD measurements, illustrative examples, and full derivations.

\subsection{Autoregressive decoding and token readiness.}
Consider a decoder-only language model with $L$ layers that generates $T$ new tokens. Standard autoregressive (AR) decoding will have $LT$ layer computations. However, for many tokens, an intermediate layer may already produce a candidate that agrees with the final-depth decoding outcome. We use this agreement to define when a token becomes \emph{ready} along the depth axis.

\begin{definition}[Acceptance and stable earliest available depth]
Let $[L]=\{1,\ldots,L\}$ and $[T]=\{1,\ldots,T\}$. For token position $t\in[T]$ and depth $\ell\in[L]$, let
$\Omega_{\mathcal R}(t,\ell)\in\{0,1\}$ denote whether the candidate token produced at depth $\ell$ is accepted by the final-depth decoding rule $\mathcal R$. We set $\Omega_{\mathcal R}(t,L)=1$ by construction. The earliest available depth is
\begin{equation}
    \mathrm{EAD}_{\mathcal R}(t)
    =\min\{\ell\in[L]:\Omega_{\mathcal R}(t,\ell)=1\}.
\end{equation}
Because intermediate predictions can occasionally match and then diverge at later layers, we use a stable version as the default quantity. Define stable acceptance by
\begin{equation}
    \Omega_{\mathcal R}^{\mathrm{stab}}(t,\ell)=1
    \quad\Longleftrightarrow\quad
    \Omega_{\mathcal R}(t,n)=1,\;\forall n\in\{\ell,\ell+1,\ldots,L\}.
\end{equation}
The stable earliest available depth is
\begin{equation}
    \mathrm{EAD}^{\mathrm{stab}}_{\mathcal R}(t)
    =\min\{\ell\in[L]:\Omega_{\mathcal R}^{\mathrm{stab}}(t,\ell)=1\}.
\end{equation}
Unless otherwise stated, EAD refers to this stable EAD and we omit the subscript $\mathcal R$ when the decoding rule is fixed.
\end{definition}

\subsection{Depth selection and its bottleneck.}
Existing lossless depth-adaptive methods typically instantiate depth adaptation as a selection problem: they choose one non-final depth, draft a token from that depth, and rely on final-depth verification to preserve equivalence with AR decoding.

\begin{definition}[Depth selection]
A depth-selection policy assigns a selected non-final exit depth $s_t\in[L-1]$ to each drafted token position $t$. Under the stable-readiness convention used in this analysis, the selected exit can exploit token readiness only when $\Omega^{\mathrm{stab}}(t,s_t)=1$, equivalently $\mathrm{EAD}(t)\le s_t$.
\end{definition}

Let $p_t^{\mathrm{acc}}$ denote the probability that the selected exit at $s_t$ is accepted under this workload model. If this exit fails, final-depth verification must replace the token, and any drafted descendants that depend on the rejected token must be discarded. Let $v_t\ge1$ be the number of discarded selected-exit tokens from position $t$ onward, including token $t$. The expected layer cost charged to position $t$ is
\begin{equation}
    W_t^{\mathrm{sel}}
    =p_t^{\mathrm{acc}}s_t
    +(1-p_t^{\mathrm{acc}})\left(L+\sum_{i=t}^{t+v_t-1}s_i\right).
\end{equation}

\begin{proposition}[Selection overhead decomposition]
\label{prop:main-selection-overhead}
Relative to an oracle that directly uses $\mathrm{EAD}(t)$, the expected overhead of depth selection is
\begin{equation}
    c_t^{\mathrm{sel}}
    =W_t^{\mathrm{sel}}-\mathrm{EAD}(t)
    =p_t^{\mathrm{acc}}(s_t-\mathrm{EAD}(t))
    +(1-p_t^{\mathrm{acc}})\left(L+\sum_{i=t}^{t+v_t-1}s_i-\mathrm{EAD}(t)\right).
\end{equation}
\end{proposition}

The derivation and a numerical example are provided in Appendix~\ref{app:selection-proof}. Proposition~\ref{prop:main-selection-overhead} exposes the selection bottleneck of this paradigm. 

\subsection{Depth exploration.}
We next formulate depth exploration as exposing a finite set of candidate depths and charging each token to the earliest explored depth that reaches stable readiness.

\begin{definition}[Exploration set and resolution]
\label{def:main-exploration-set}
Let $\mathcal X=\{d_0<d_1<\cdots<d_{N-1}=L\}\subseteq[L]$ be the explored depth set. The final depth is always included so that full-depth decoding remains available. For any $\ell\in[L]$, define the depth ceiling
\begin{equation}
    \lceil \ell\rceil_{\mathcal X}=\min\{d\in\mathcal X:d\ge \ell\}.
\end{equation}
The resolution of $\mathcal X$ is
\begin{equation}
    \Delta(\mathcal X)=\max_{\ell\in[L]}\left(\lceil \ell\rceil_{\mathcal X}-\ell\right),
\end{equation}
which measures the largest upward rounding error caused by restricting usable depths to $\mathcal X$.
\end{definition}

\begin{definition}[Depth exploration]
\label{def:main-depth-exploration}
Given $\mathcal X$, the stable depth-exploration workload for token $t$ is the earliest explored depth that reaches the stable-readiness threshold:
\begin{equation}
    W_{\mathcal X}(t)
    =\min\{d\in\mathcal X:\Omega^{\mathrm{stab}}(t,d)=1\}.
\end{equation}
Equivalently,
\begin{equation}
    W_{\mathcal X}(t)=\lceil \mathrm{EAD}(t)\rceil_{\mathcal X}.
\end{equation}
\end{definition}

\begin{proposition}[Resolution-bounded exploration overhead]
\label{prop:main-exploration-overhead}
For any token $t$ and exploration set $\mathcal X$, the overhead of depth exploration relative to the EAD oracle is
\begin{equation}
    c_t^{\mathrm{exp}}
    =W_{\mathcal X}(t)-\mathrm{EAD}(t)
    =\lceil \mathrm{EAD}(t)\rceil_{\mathcal X}-\mathrm{EAD}(t),
\end{equation}
so
\begin{equation}
    0\le c_t^{\mathrm{exp}}\le\Delta(\mathcal X).
\end{equation}
\end{proposition}

A proof and an example for uniformly spaced exploration sets are provided in Appendix~\ref{app:exploration-proof}.

\noindent\textbf{Depth-side speedup.}
We use the same layer-cost accounting to relate these depth quantities to theoretical speedup. This accounting ignores non-layer costs and assumes uniform per-layer cost; the experiments in Section~4 report measured walltime throughput.

\begin{proposition}[Depth-side speedup and monotonicity]
\label{prop:main-speedup-monotonicity}
For any decoding algorithm with charged layer cost $W(t)$ for token $t$, its depth-side speedup over AR decoding is
\begin{equation}
\label{eq:main-speedup-general}
    S=\frac{LT}{\sum_{t=1}^{T}W(t)}.
\end{equation}
For depth exploration,
\begin{equation}
\label{eq:main-speedup-exploration}
    S_{\mathcal X}
    =\frac{LT}{\sum_{t=1}^{T}W_{\mathcal X}(t)}
    =\frac{LT}{\sum_{t=1}^{T}\lceil\mathrm{EAD}(t)\rceil_{\mathcal X}}.
\end{equation}
By Proposition~\ref{prop:main-exploration-overhead},
\begin{equation}
\label{eq:main-speedup-bound}
    1\le
    \frac{LT}{\sum_{t=1}^{T}\min\{\mathrm{EAD}(t)+\Delta(\mathcal X),L\}}
    \le S_{\mathcal X}
    \le
    \frac{LT}{\sum_{t=1}^{T}\mathrm{EAD}(t)}
    =S_{\mathrm{EAD}}.
\end{equation}
Here $S_{\mathrm{EAD}}$ is the oracle speedup obtained if every token's EAD were known and directly usable. Moreover, if $\mathcal X\subseteq\mathcal X'$, then $S_{\mathcal X}\le S_{\mathcal X'}\le S_{\mathrm{EAD}}$; finer exploration monotonically approaches the EAD oracle under this depth-side accounting.
\end{proposition}

The derivation of Eq.~\eqref{eq:main-speedup-bound} and the monotonicity statement is provided in Appendix~\ref{app:speedup-proof}. While this analysis characterizes the ideal depth-side opportunity of exploration, Section~\ref{sec:lattice_fsm} revisits this under a more realistic setting. The next section instantiates this idealized exploration model as an exact decoding algorithm and discusses the additional walltime costs introduced by branch execution.
\section{Method}
\label{sec:method}

\subsection{Algorithm overview}

 We now describe the \sys algorithm. Conceptually, \sys~maps each depth stage \(\Pi_k=\mathrm{Layers}(d_{k-1},d_k]\) to a depth explorer \(E_k\), with each explorer deployed on a separate hardware computing unit. The output boundary of \(E_k\) corresponds to candidate depth \(d_k\in\mathcal X\). Each explorer processes one depth stage and exposes candidates at its boundary. DEX commits the final-depth token, selects the earliest matching branch for reuse, and collapses stale branches. This resembles the principle of carry-lookahead and Kogge-Stone adders~\cite{weinberger1958logic,kogge1973parallel}, where parallel candidate computations expose a sequential dependency earlier than a purely serial execution. The whole execution procedure is demonstrated in Algorithm~\ref{alg:dex}.

\begin{algorithm}[t]
\caption{Depth Exploration Decoding (\sys)}
\label{alg:dex}
\begin{algorithmic}[1]
\Require Prompt tokens \(\mathbf{x}\); exploration set
\(\mathcal X=\{d_0,\ldots,d_{N-1}\}\), where \(d_{N-1}\) is the final depth;
stages \(\Pi_k=\mathrm{Layers}(d_{k-1},d_k]\) with \(d_{-1}=0\); max token length \(T\)
\Ensure Generated tokens \(Y\)

\State \(E_k \gets (\Pi_k,\mathrm{KV}_k)\), for \(k=0,\ldots,N-1\)
\Comment{depth explorer}
\State \(Y\gets[\,]\), \(B\gets\emptyset\), \(q\gets0\)
\Comment{\(B\): explore buffer; \(q\): FSM state}
\State \(S\gets\fn{init}(\mathbf{x},\{E_k\})\)
\Comment{\(S\): hidden states and per-stage KV views}

\While{\(|Y|<T\)}
    \State \(H_q\gets\fn{expand}(\{E_k\},S,q)\)
    \Comment{parallel lattice expansion; Section~\ref{sec:lattice_fsm}}

    \State \(Z_q\gets\fn{DECS}(E_0,H_q,B,q)\)
    \Comment{depth-coupled sampling; Section~\ref{sec:decs}}

    \State \(B\gets\fn{insert}(B,Z_q)\)

    \If{\(\fn{ReachFinalDepth}(q)\)}
        \State \((b,y)\gets\fn{validate\_and\_commit}(B)\)
        \Comment{\(b\): accepted branch; \(y\): committed token}
        \State \(Y\gets Y\Vert y\)

        \If{\(\fn{stop}(Y)\)}
            \State \Return \(Y\)
        \EndIf

        \State \((B,S,q)\gets\fn{collapse}(B,S,q,b)\)
        \Comment{post-commit lattice collapse; Section~\ref{sec:lattice_fsm}}
    \Else
        \State \(q\gets\fn{advance}(q)\)
        \Comment{advance FSM state; Section~\ref{sec:lattice_fsm}}
    \EndIf
\EndWhile

\State \Return \(Y\)
\end{algorithmic}
\end{algorithm}

\subsection{Lattice expansion and commit--collapse}
\label{sec:lattice_fsm}

\begin{figure}[t]
    \centering
    \includegraphics[width=\linewidth]{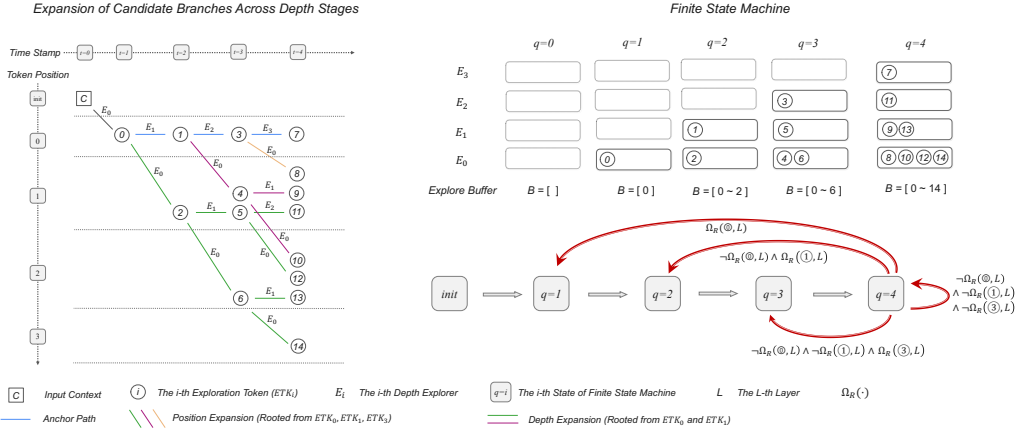}

    \caption{
    % \TODO{fix figure error cell number wrong}
\sys execution with four depth explorers.
(Left) Candidate branches are unrolled on a depth--position lattice, where the anchor path reaches the final depth while descendant branches are precomputed for potential reuse.
(Right) The FSM records the explorer round, materialized buffer entries, and collapse transition after the final-depth reference determines the committed token.
}
    \label{fig:lattice_fsm}
\end{figure}

\noindent\textbf{Lattice expansion.}
To correctly manage candidate sequences generated by different depth explorers, we organize each exploration cycle around a \emph{commit position} \(p_c\), i.e., the next token position to be appended to the output sequence. In Figure~\ref{fig:lattice_fsm} (left), this is the row marked as position 0. We represent depth exploration as a depth--position lattice unrolled over explorer rounds, using four explorers \(E_0\sim E_3\), each covering one quarter of the model depth. Each circle denotes an exploration token \(\mathrm{ETK}_i\), where \(i\) is its token id in the explore buffer.

At each round, an exploration token can expand in two directions. A depth expansion keeps the same token position and forwards its hidden state to the next explorer. A position expansion uses the current candidate token as the prefix for the next token position and starts that position from \(E_0\). Each position expansion creates a new \emph{sampling slot}, i.e., a branch-conditioned next-token determined by its prefix history and target token position. The ETKs that instantiate the same sampling slot at different depths form a \emph{slot chain}. The slot chain for the current commit position under the committed context is the \emph{anchor path}. For example, \(E_0\) first produces \(\mathrm{ETK}_0\) for \(p_c\); in the next round, \(\mathrm{ETK}_0\)'s hidden state continues through \(E_1\) to produce \(\mathrm{ETK}_1\) in the same slot chain, while \(\mathrm{ETK}_0\) is also used as a prefix to create a new slot whose first token is \(\mathrm{ETK}_2\).

After four explorer rounds, the anchor path reaches the final-depth node, which provides the reference token for the commit position. Other slot chains correspond to speculative descendants under branch-specific prefixes. Before collapse, the lattice follows a binary expansion structure, allowing ancestry, KV entries, and attention masks to be indexed consistently.

\noindent\textbf{Commit process.}
Once the anchor path reaches the final depth, \sys~uses its final-depth token as the commit reference. Let \((b_0,u_0),\ldots,(b_{N-1},u_{N-1})\) denote the anchor-path branches and their candidate tokens at \(p_c\), ordered by increasing depth, where \(u_{N-1}\) is the final-depth reference. \sys~checks whether \(u_{N-1}\) appears among the non-final anchor candidates \(\{u_0,\ldots,u_{N-2}\}\); if so, it selects the shallowest matching branch \(b_k\), and otherwise falls back to the final-depth branch \(b_{N-1}\). In both cases, the token appended to the official output is \(y=u_{N-1}\), while the selected branch \(b\) determines which part of the lattice can be preserved during collapse.

\noindent\textbf{Finite-state control and collapse.}
\sys~employs a finite-state machine (FSM) to govern branch lifecycles during depth exploration, as shown in Figure~\ref{fig:lattice_fsm} (right). A state \(q\) records how many explorer rounds have been completed since the last committed token and determines whether the system should continue expansion or perform commitment. For each state, the figure shows the ETKs produced after forward execution, the resulting explore buffer \(B\), and the per-explorer KV-cache views, where \(C\) denotes committed context tokens. Under the four-explorer setup, state \(q\) materializes the first \(2^q-1\) tokens in the binary expansion tree, e.g., \(B=\{\mathrm{ETK}_0,\ldots,\mathrm{ETK}_{14}\}\) at state 4.

\noindent\textbf{Lossless invariant.}
\sys is lossless with respect to standard autoregressive decoding because commitment is always defined by the final-depth reference. At every commit position, the committed prefix and the final-depth KV state maintained by \sys are identical to those of standard autoregressive decoding under the same decoding rule. The final-depth anchor node is computed from this committed prefix using the unmodified backbone model. \sys appends exactly the token produced by this node: if an earlier anchor candidate matches the final-depth token, the earlier branch only determines how much of the lattice can be reused; the committed token itself remains the final-depth token. During collapse, \sys preserves only nodes whose prefix equals the new committed prefix and whose per-stage hidden states and KV entries were computed under that prefix; all inconsistent nodes are discarded or recomputed. Therefore, by induction over commit positions, \sys produces the same token sequence and maintains the same final-depth state as standard autoregressive decoding. For sampling, the final-depth reference is sampled from unmasked final-depth logits using the target random source, so the committed sample distribution is unchanged.

Proposition~\ref{prop:main-speedup-monotonicity} captures the ideal depth-side opportunity. In practice, DEX also pays branch-execution, communication, KV-management, and collapse overheads; Appendix~\ref{app:overhead} profiles these costs and shows that commit–collapse overhead is small relative to expansion.

\subsection{Optimized depth-coupled sampling}
\label{sec:decs}

If candidates in each slot chain are generated independently, different depths may repeatedly propose the same token. Such duplicates are unhelpful for \sys: if a shallower proposal \(v\) already matches the final-depth reference, the shallower branch would be selected; if it does not match, repeating \(v\) only reduces the opportunity to expose another candidate that may hit the reference.

To reduce duplicate proposals, \sys~uses Depth-Coupled Sampling (DECS). 
Within each slot chain, DECS masks tokens already proposed by shallower depths when decoding later non-final candidates, encouraging deeper explorers to cover alternatives. 
The final-depth reference is never masked; in sampling mode, it is decoded from unmasked logits with the slot-chain Gumbel source, so the Gumbel-Max trick~\cite{jang2017categorical} preserves the exact target sample. 
DECS therefore only diversifies intermediate proposals while preserving lossless commitment. 
The full algorithm is given in Appendix~\ref{app:decs}.

\subsection{Adapt to standard LLM with inducing adapter}
\label{sec:adapter}

By Proposition~\ref{prop:main-speedup-monotonicity}, the speedup of \sys is governed by the EAD distribution. Therefore, \sys can be directly applied to EAD-favorable LLMs, where most tokens already have shallow EADs. In standard LLMs, however, shallow and middle-layer hidden states are often not well aligned with the shared LM head, causing stable EADs to concentrate near the final depth. Motivated by prior work showing that early-exit behavior can be induced through pretraining or parameter-efficient tuning~\cite{chen2024eellm,elhoushi2024layerskip,wei2025adadecode}, we attach small residual adapters to these middle candidate depths to align intermediate hidden states with the shared LM head. These adapters are trained by self-distillation from the frozen final-depth model and affect only intermediate proposals; final-depth commitment remains unchanged. Details are in Appendix~\ref{app:adapter}.
\section{Experiment}
\label{sec:exp}

\noindent\textbf{LLMs and datasets.}
We evaluate \sys on both early-exit-trained and standard LLMs. For depth-axis analysis, we use two LayerSkip-trained models, CodeLlama-34B and Llama-2-70B, denoted as LS-CodeLlama-34B and LS-Llama-2-70B. These models expose intermediate-layer predictions and are therefore suitable for measuring EAD, oracle headroom, and the gap between \sys and depth-selection baselines. We also evaluate standard CodeLlama-34B-Instruct~\cite{roziere2024codellamaopenfoundation}, Llama-2-70B-Instruct~\cite{touvron2023llama}, and Qwen3-32B~\cite{yang2025qwen3technicalreport} to study \sys's end-to-end throughput under regular model settings and compare it with broader decoding accelerators. We use GSM8K for mathematical reasoning~\cite{cobbe2021gsm8k}, HumanEval for code generation~\cite{chen2021evaluating}, and XSum for summarization~\cite{Narayan2018DontGM}.

\noindent\textbf{Baselines and setup.}
We organize baselines into two groups. First, for depth-axis comparison, we compare \sys with representative depth-selection methods: LayerSkip self-speculative decoding (LSSD)~\cite{elhoushi2024layerskip}, AdaDecode~\cite{wei2025adadecode}, and DEL~\cite{zarch2025del}. Second, for end-to-end throughput comparison, we include autoregressive decoding (AR), tensor-parallel decoding (TP), Lookahead~\cite{fu2024break}, AdaDecode~\cite{wei2025adadecode}, PEARL~\cite{liu2025pearl}, and the EAGLE series~\cite{li2024eagle2,li2025eagle3}. All methods are evaluated on NVIDIA H100 GPUs. TP is tested with tensor-parallel sizes 2, 4, and 8, and we report its best throughput. LSSD/AdaDecode/DEL/Lookahead/PEARL/EAGLE use 3 GPUs for 32B/34B models and 4 GPUs for 70B models. \sys is evaluated under multiple uniformly partitioned exploration resolutions, denoted as DEX-\(1/K\), where \(K\) is the number of depth explorers, each running on one GPU. For example, DEX-\(1/3\) uses three explorers and adopts a granularity of one third of the full model depth. Unless specified, we use greedy decoding with batch size 1 and maximum generation length 512 for Llama-family models and 1024 for Qwen3-32B. Additional details are provided in Appendix~\ref{app:eval}.

\subsection{Walltime comparison with depth-selection methods}

\begin{figure}[t]
    \centering
    \includegraphics[width=0.48\linewidth]{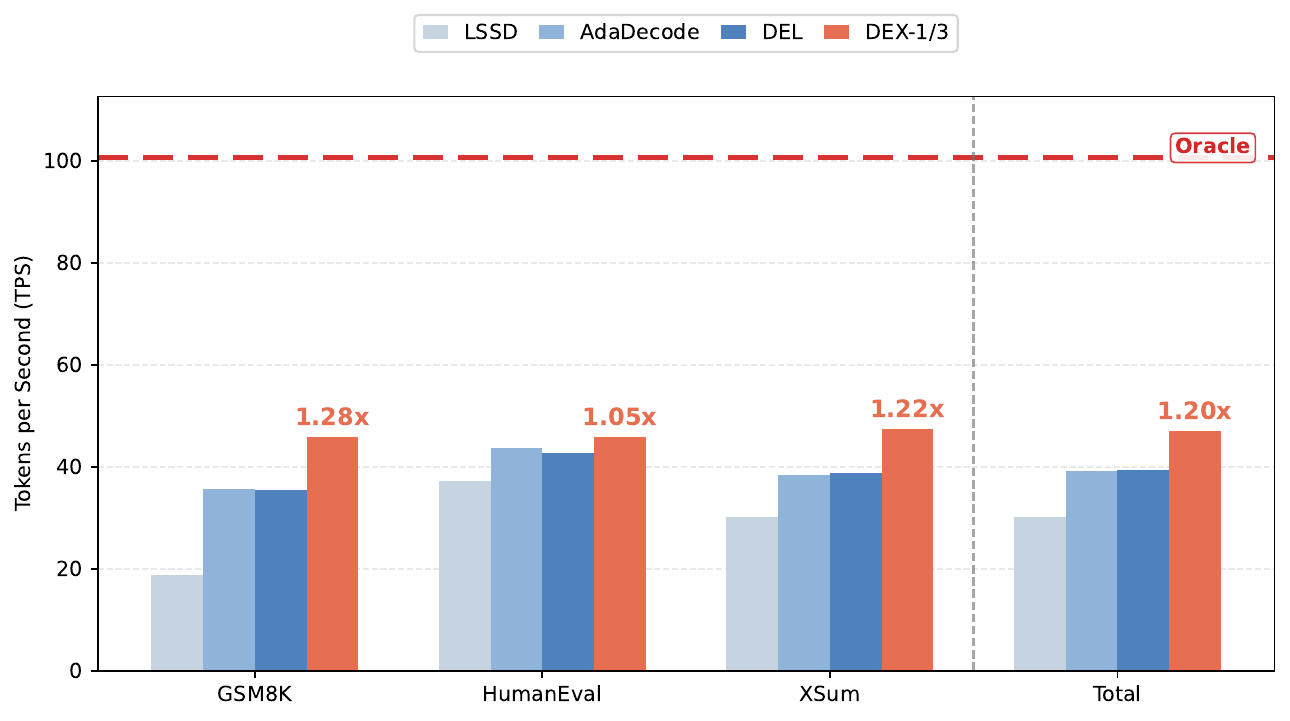}
    \hfill
    \includegraphics[width=0.48\linewidth]{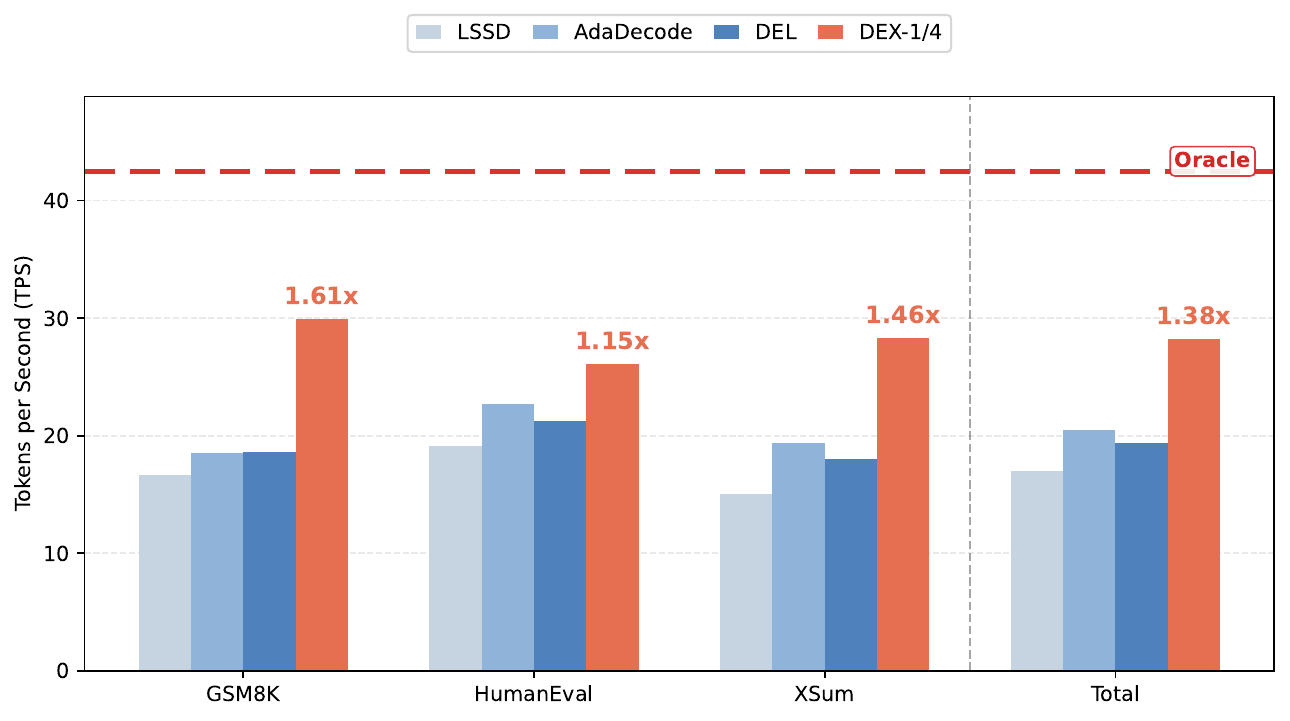}

    \caption{
    Depth-axis throughput comparison on LS-CodeLlama-34B (left) and LS-Llama-2-70B (right).
    \sys outperforms depth-selection baselines across datasets, and the dashed line marks the oracle throughput implied by EAD headroom.
    }
    \label{fig:depth_speed}
\end{figure}

\noindent\textbf{Throughput comparison.}
Figure~\ref{fig:depth_speed} compares \sys~with representative depth-selection methods on LayerSkip-trained models, where non-final-layer predictions are explicitly exposed. \sys~achieves higher measured throughput than LayerSkip, AdaDecode, and DEL across both models and all datasets. The dashed oracle line denotes the throughput implied by stable EAD measurements, serving as a depth-side headroom reference rather than an implemented baseline. The remaining oracle gap indicates further depth-side headroom, but \sys realizes a larger fraction of it than the selection baselines.

\begin{figure}[t]
    \centering
    % left
    \begin{minipage}[c]{0.58\linewidth}
        \centering
        \includegraphics[
            width=\linewidth
        ]{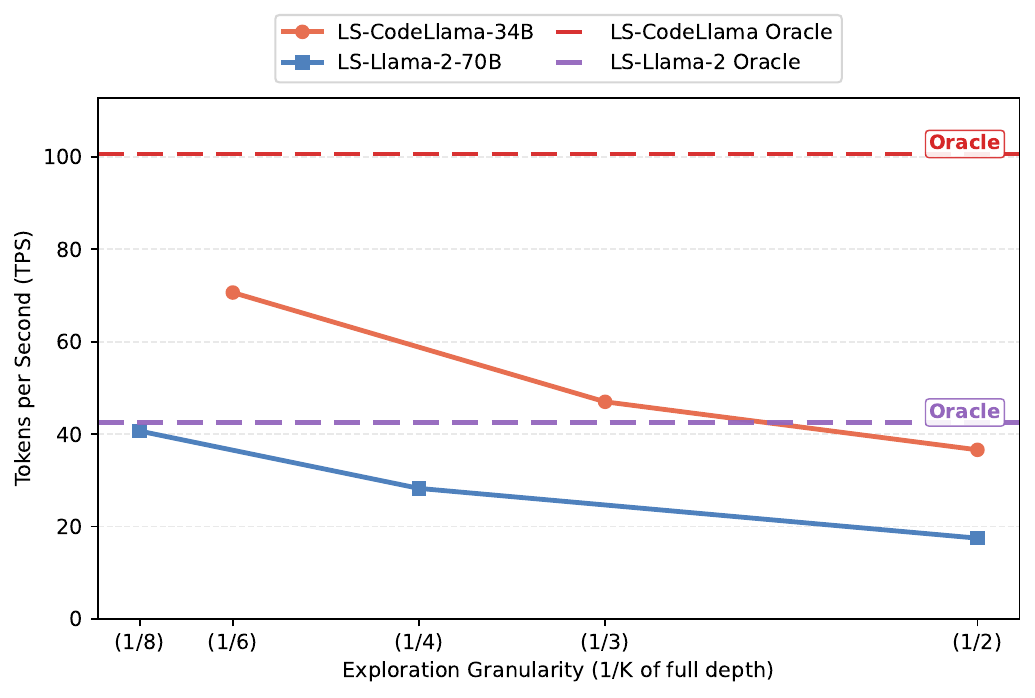}
    \end{minipage}
    \hfill
    % right
    \begin{minipage}[c]{0.38\linewidth}
        \centering
        \includegraphics[
            width=\linewidth
        ]{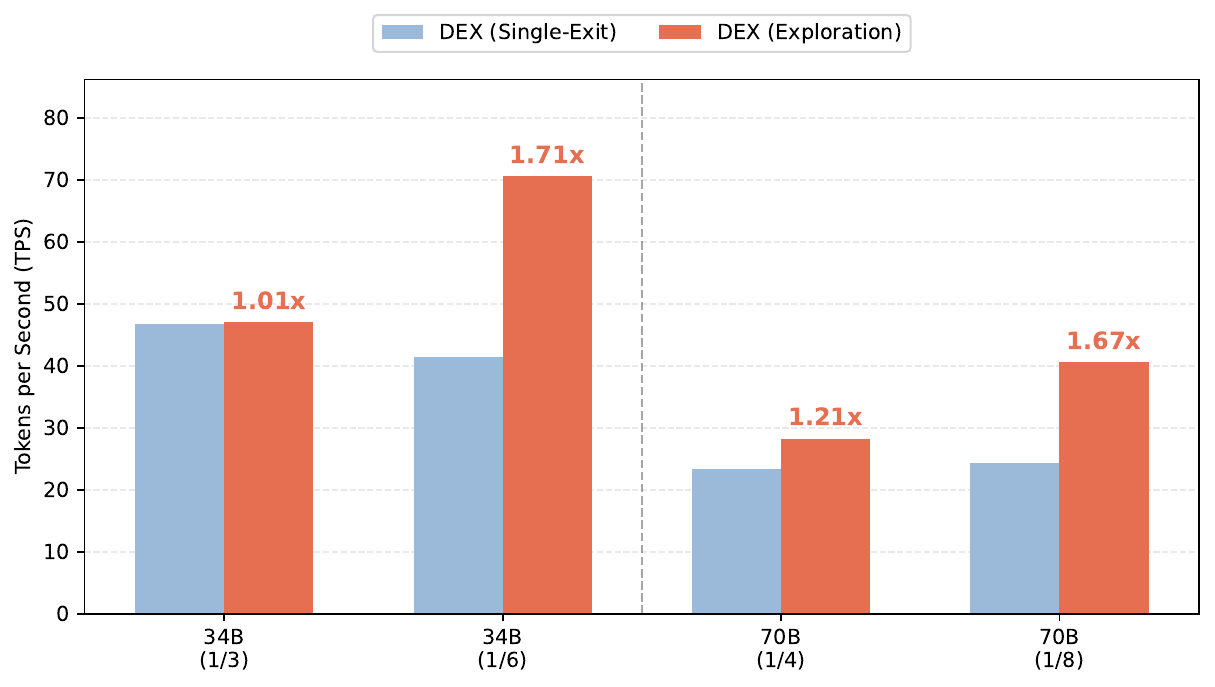}
        \vspace{0.6em}
        \includegraphics[
            width=\linewidth
        ]{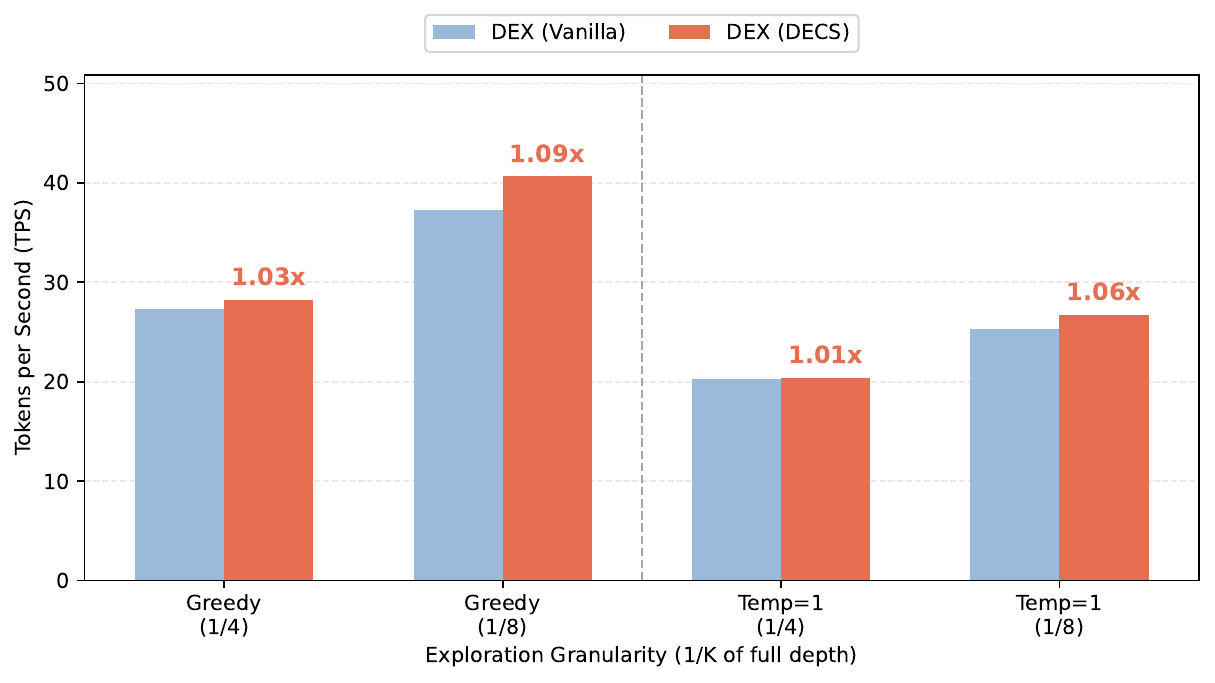}
    \end{minipage}
    \caption{
    Scaling behavior and architectural analysis of \sys.
    Left: throughput under different exploration resolutions.
    Right-top: comparison between single-exit decoding and depth exploration.
    Right-bottom: comparison between vanilla decoding and DECS-enhanced exploration in both greedy and sampling (Temperature=1.0) cases.
    }
    \label{fig:scaling_analysis}
\end{figure}

\noindent\textbf{Scaling with exploration resolution.}
Figure~\ref{fig:scaling_analysis} (left) shows that finer exploration generally improves throughput. This trend is consistent with the formulation in Section~\ref{sec:prelim}: increasing the number of explored depths lets \sys~track token readiness with a smaller upward rounding error. The gain is especially clear on LS-Llama-2-70B, where increasing the exploration resolution substantially narrows the gap to the EAD-implied oracle. This indicates that exploration resolution acts as an effective control knob for converting additional depth explorers into realized throughput.

\noindent\textbf{Ablation study.}
Figure~\ref{fig:scaling_analysis} (right-top) isolates the effect of exploration by comparing \sys~with a single-exit variant that disables intermediate depth branches except for the first and final depth candidates. At coarse resolutions, the two variants are close; at finer resolutions, full exploration gives a much larger gain. This suggests that the benefit comes not merely from running depth stages in parallel, but from exposing multiple candidate depths and committing the earliest branch accepted by the final-depth reference. 

Figure~\ref{fig:scaling_analysis} (right-bottom) evaluates the throughput effect of DECS. Compared with vanilla decoding, DECS yields a consistent throughput improvement under both greedy decoding and temperature sampling, showing that this by-construction diversification translates into practical decoding speedups.

\subsection{End-to-end walltime comparison results}
\label{sec:e2e_compare}
We next evaluate \sys~on standard LLMs and compare it with broader decoding methods. This setting tests whether the gains observed against depth-selection methods persist beyond early-exit-trained models. As mentioned in our setup, TP and PEARL are the multi-GPU parallel baseline, and Lookahead, PEARL, and EAGLE represent strong algorithms under their official configurations.

\begin{figure}[t]
    \centering
    \begin{minipage}[c]{0.32\linewidth}
        \centering
        \includegraphics[
            width=\linewidth
        ]{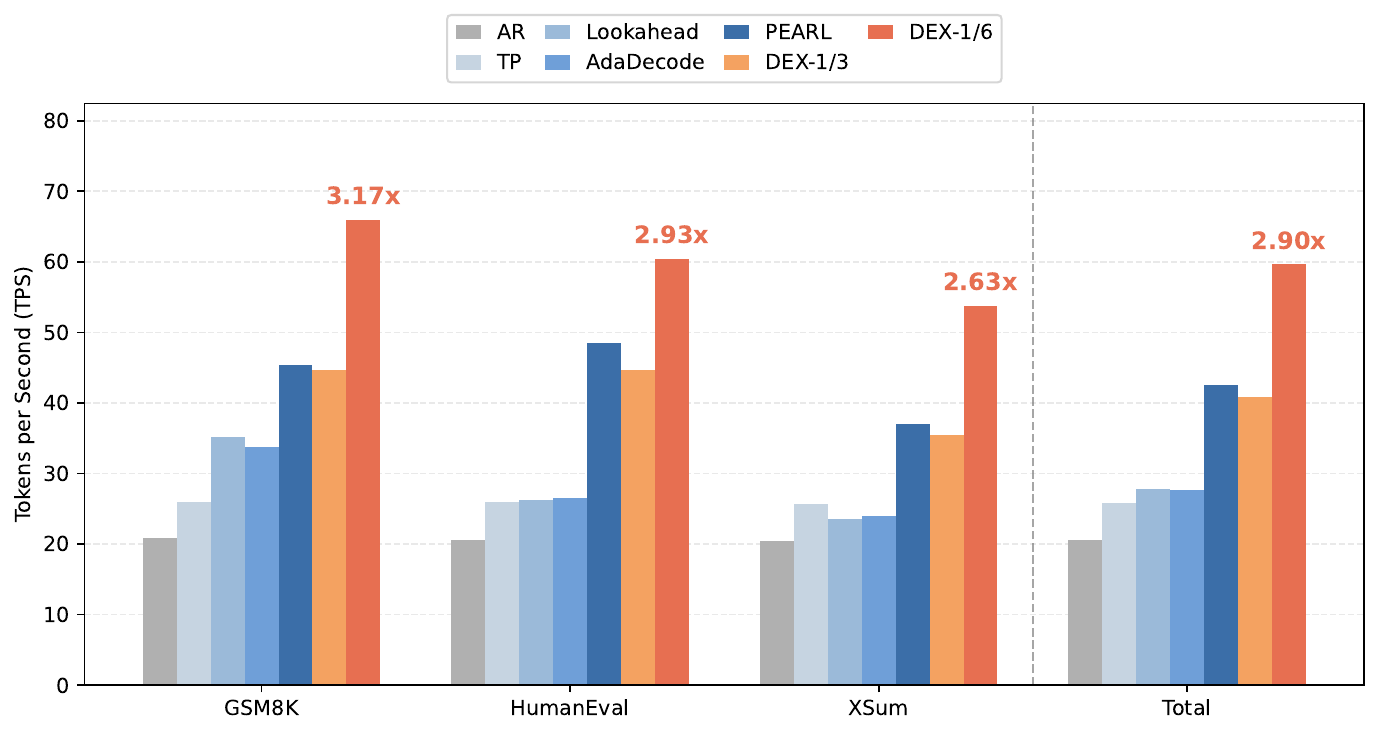}
    \end{minipage}
    \hfill
    \begin{minipage}[c]{0.32\linewidth}
        \centering
        \includegraphics[
            width=\linewidth
        ]{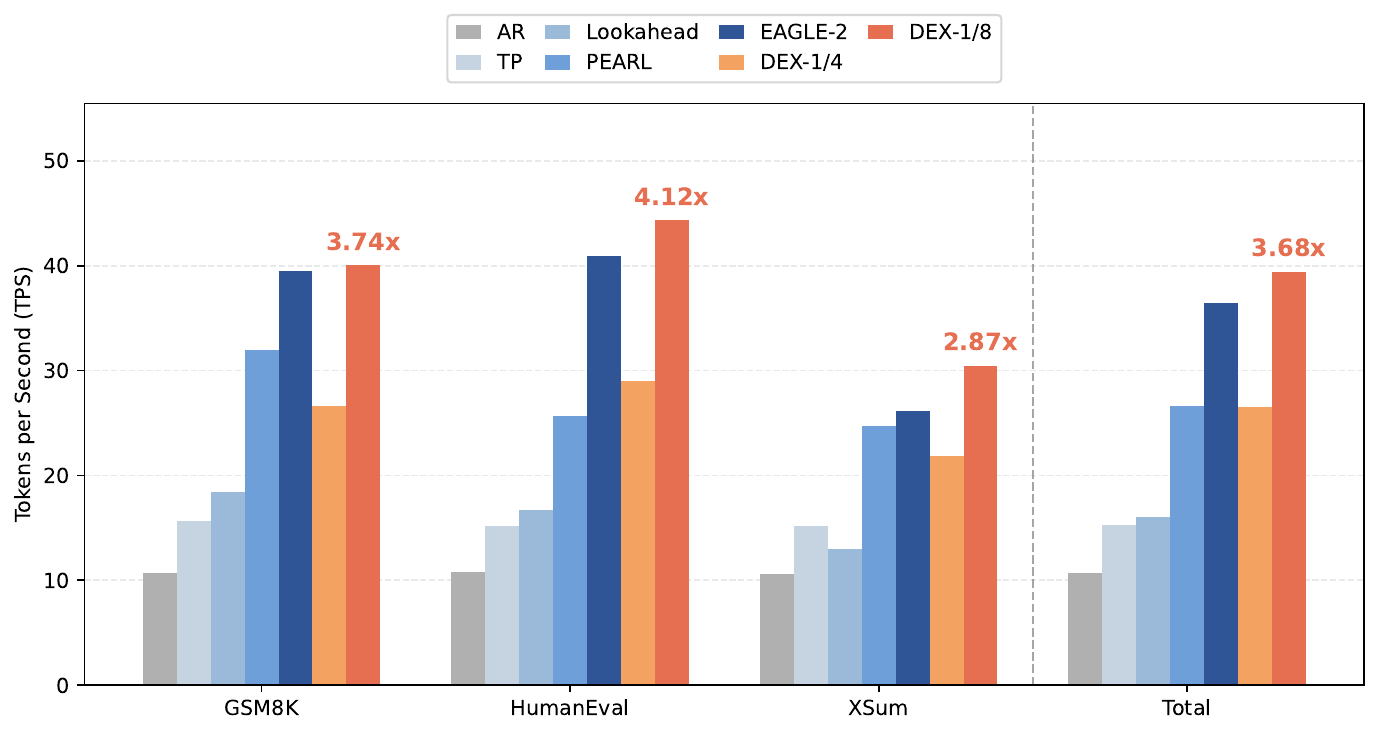}
    \end{minipage}
    \hfill
    \begin{minipage}[c]{0.32\linewidth}
        \centering
        \includegraphics[
            width=\linewidth
        ]{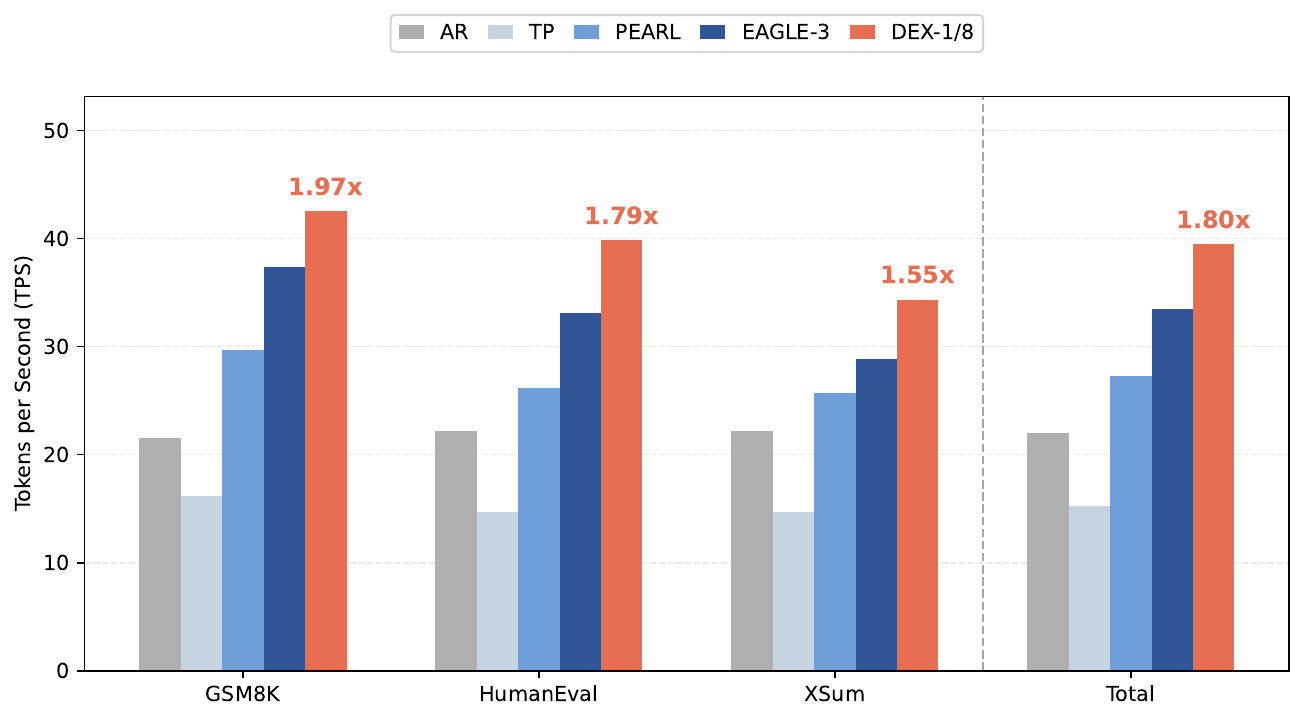}
    \end{minipage}
    \caption{
    End-to-end decoding throughput comparison across CodeLlama-34B, Llama-2-70B, and Qwen3-32B.
    Scaled \sys achieves the highest throughput among speculative and parallel decoding baselines across reasoning, coding, and summarization workloads.
    }
    \label{fig:end_to_end_throughput}
\end{figure}

Figure~\ref{fig:end_to_end_throughput} shows that \sys~achieves strong end-to-end throughput across CodeLlama-34B, Llama-2-70B, and Qwen3-32B. Under matched GPU resources, \sys~is already competitive with distributed baselines such as TP and PEARL: on CodeLlama-34B and Llama-2-70B, DEX-\(1/3\) and DEX-\(1/4\) are competitive with the matched 3-GPU and 4-GPU PEARL settings, respectively. At finer exploration resolutions, \sys~further improves throughput, with DEX-\(1/8\) achieving the highest throughput among the evaluated configurations on Llama-2-70B. On Qwen3-32B, \sys~also improves over AR, TP, PEARL, and EAGLE-3, suggesting that depth exploration remains effective on a recent LLM.

These results show that DEX is competitive under matched resources and improves further as additional explorers increase the exploration resolution. Appendix~\ref{app:discussion} provides a closer look at the resource-to-resolution tradeoff behind this scaling.

\section{Related work}

\noindent\textbf{Depth-axis LLM acceleration.}
The depth axis asks how early the next token becomes determined before the final layer.
Mixture-of-Depths exploits this redundancy through learned layer routing, but changes the model's conditional-compute architecture rather than providing a lossless wrapper for an existing full-depth decoder
\cite{raposo2024mixtureofdepthsdynamicallyallocatingcompute}.
For decoding, LayerSkip trains intermediate layers for early prediction, while Draft \& Verify, SWIFT, and PPSD-style methods use skipped, partial-depth, or pipelined executions as self-drafters followed by final-depth verification
\cite{elhoushi2024layerskip, DBLP:conf/acl/Zhang00S0CM24, xia2025swiftontheflyselfspeculativedecoding, li2026optimized}.
AdaDecode and DEL further make this path adaptive, by overlapping intermediate-layer prediction with deferred deeper computation or by adjusting the exit layer and speculation length
\cite{wei2025adadecode, zarch2025del}.
These methods still try one chosen early-exit, skip, or pipeline path in a speculative round.
However, our proposed \sys instead exposes multiple candidate depths for the same commit position and reuses the shallowest branch certified by the final-depth reference.

\noindent\textbf{Token-axis LLM acceleration.}
The token axis asks how many future token positions can be resolved from one target-model verification.
Speculative decoding uses a draft model to propose a block of tokens and the target model to verify it
\cite{leviathan2023fast, chen2023acceleratinglargelanguagemodel, wu2025easyspeclayerparallelspeculativedecoding}.
Medusa and the EAGLE family improve the draft side with extra prediction heads or feature-level drafts
\cite{cai2024medusasimplellminference, li2025eaglespeculativesamplingrequires, li2024eagle2, li2025eagle3}.
Lookahead Decoding avoids a separate draft model by using the target model itself to find and verify short candidate continuations
\cite{fu2024break}.
PEARL and Speculative Speculative Decoding focus on the scheduling of the draft--verify loop, reducing idle time through adaptive draft lengths or predicted verification outcomes
\cite{liu2025pearl, kumar2026speculativespeculativedecoding}.
These methods speed up decoding by accepting more future tokens per target-model step.
In contrast, DEX speeds up the current commit position by exploring multiple depths before final-depth commitment.
The two axes are complementary in principle, but combining them would require coordinating token-tree verification with DEX's depth-lattice collapse.

\noindent\textbf{Parallel execution for LLM decoding.}
Another direction improves decoding throughput by better using parallel hardware.
Tensor and pipeline parallelism partition full-depth model execution across devices, while decoding-specific systems such as EasySpec and PPSD-style methods overlap parts of the draft--verify pipeline or parallelize layer execution
\cite{shoeybi2020megatronlmtrainingmultibillionparameter, huang2019gpipeefficienttraininggiant, wu2025easyspeclayerparallelspeculativedecoding, li2026optimized}.
DEX also uses multiple devices, but the devices are not merely shards of one full-depth pass or one draft--verify pipeline.
Instead, they act as depth explorers that expose competing candidate depths, after which the final-depth reference commits the token and collapses the exploration lattice.

\section{Conclusion}

By formalizing token readiness through EAD, we showed that selection-based methods are limited by the need to choose one exit depth under uncertainty, whereas exploration converts this uncertainty into a resolution-controlled upward rounding problem over candidate depths. Building upon that, we presented \sys, a lossless decoding algorithm that realizes this formulation through an expand--commit--collapse procedure. Across early-exit-trained and standard LLMs, \sys closes more of the depth-side headroom than prior depth-selection methods and achieves competitive end-to-end throughput against representative speculative and distributed decoding baselines, with stronger scaling as more depth explorers are available.  These results establish parallel depth exploration as a practical and scalable way for exploiting the underused depth axis of autoregressive LLM decoding.

\bibliographystyle{unsrtnat}
\bibliography{paper}

@inproceedings{elhoushi2024layerskip,
    title = "{L}ayer{S}kip: Enabling Early Exit Inference and Self-Speculative Decoding",
    author = "Elhoushi, Mostafa  and
      Shrivastava, Akshat  and
      Liskovich, Diana  and
      Hosmer, Basil  and
      Wasti, Bram  and
      Lai, Liangzhen  and
      Mahmoud, Anas  and
      Acun, Bilge  and
      Agarwal, Saurabh  and
      Roman, Ahmed  and
      Aly, Ahmed  and
      Chen, Beidi  and
      Wu, Carole-Jean",
    editor = "Ku, Lun-Wei  and
      Martins, Andre  and
      Srikumar, Vivek",
    booktitle = "Proceedings of the 62nd Annual Meeting of the Association for Computational Linguistics (Volume 1: Long Papers)",
    month = aug,
    year = "2024",
    address = "Bangkok, Thailand",
    publisher = "Association for Computational Linguistics",
    url = "https://aclanthology.org/2024.acl-long.681/",
    doi = "10.18653/v1/2024.acl-long.681",
    pages = "12622--12642"
}

@article{touvron2023llama,
  title={Llama 2: Open foundation and fine-tuned chat models},
  author={Touvron, Hugo and Martin, Louis and Stone, Kevin and Albert, Peter and Almahairi, Amjad and Babaei, Yasmine and Bashlykov, Nikolay and Batra, Soumya and Bhargava, Prajjwal and Bhosale, Shruti and others},
  journal={arXiv preprint arXiv:2307.09288},
  year={2023}
}

@article{kogge1973parallel,
  title={A parallel algorithm for the efficient solution of a general class of recurrence equations},
  author={Kogge, Peter M and Stone, Harold S},
  journal={IEEE transactions on computers},
  volume={100},
  number={8},
  pages={786--793},
  year={1973},
  publisher={IEEE}
}

@article{weinberger1958logic,
  author  = {Weinberger, A. and Smith, J. L.},
  title   = {A Logic for High-Speed Addition},
  journal = {National Bureau of Standards Circular},
  volume  = {591},
  pages   = {3--12},
  year    = {1958}
}

@inproceedings{
jang2017categorical,
title={Categorical Reparameterization with Gumbel-Softmax},
author={Eric Jang and Shixiang Gu and Ben Poole},
booktitle={International Conference on Learning Representations},
year={2017},
url={https://openreview.net/forum?id=rkE3y85ee}
}

@article{wei2025adadecode,
  title={Adadecode: Accelerating llm decoding with adaptive layer parallelism},
  author={Wei, Zhepei and Chen, Wei-Lin and Zhu, Xinyu and Meng, Yu},
  journal={arXiv preprint arXiv:2506.03700},
  year={2025}
}

@article{Narayan2018DontGM,
  title={Don't Give Me the Details, Just the Summary! Topic-Aware Convolutional Neural Networks for Extreme Summarization},
  author={Shashi Narayan and Shay B. Cohen and Mirella Lapata},
  journal={ArXiv},
  year={2018},
  volume={abs/1808.08745}
}

@misc{chen2021evaluating,
      title={Evaluating Large Language Models Trained on Code},
      author={Mark Chen and Jerry Tworek and Heewoo Jun and Qiming Yuan and Henrique Ponde de Oliveira Pinto and Jared Kaplan and Harri Edwards and Yuri Burda and Nicholas Joseph and Greg Brockman and Alex Ray and Raul Puri and Gretchen Krueger and Michael Petrov and Heidy Khlaaf and Girish Sastry and Pamela Mishkin and Brooke Chan and Scott Gray and Nick Ryder and Mikhail Pavlov and Alethea Power and Lukasz Kaiser and Mohammad Bavarian and Clemens Winter and Philippe Tillet and Felipe Petroski Such and Dave Cummings and Matthias Plappert and Fotios Chantzis and Elizabeth Barnes and Ariel Herbert-Voss and William Hebgen Guss and Alex Nichol and Alex Paino and Nikolas Tezak and Jie Tang and Igor Babuschkin and Suchir Balaji and Shantanu Jain and William Saunders and Christopher Hesse and Andrew N. Carr and Jan Leike and Josh Achiam and Vedant Misra and Evan Morikawa and Alec Radford and Matthew Knight and Miles Brundage and Mira Murati and Katie Mayer and Peter Welinder and Bob McGrew and Dario Amodei and Sam McCandlish and Ilya Sutskever and Wojciech Zaremba},
      year={2021},
      eprint={2107.03374},
      archivePrefix={arXiv},
      primaryClass={cs.LG}
}

@article{cobbe2021gsm8k,
  title={Training Verifiers to Solve Math Word Problems},
  author={Cobbe, Karl and Kosaraju, Vineet and Bavarian, Mohammad and Chen, Mark and Jun, Heewoo and Kaiser, Lukasz and Plappert, Matthias and Tworek, Jerry and Hilton, Jacob and Nakano, Reiichiro and Hesse, Christopher and Schulman, John},
  journal={arXiv preprint arXiv:2110.14168},
  year={2021}
}

@inproceedings{
zarch2025del,
title={{DEL}: Context-Aware Dynamic Exit Layer for Efficient Self-Speculative Decoding},
author={Hossein Entezari Zarch and Lei Gao and Chaoyi Jiang and Murali Annavaram},
booktitle={Second Conference on Language Modeling},
year={2025},
url={https://openreview.net/forum?id=cAFxSuXQvT}
}

@inproceedings{fu2024break,
author = {Fu, Yichao and Bailis, Peter and Stoica, Ion and Zhang, Hao},
title = {Break the sequential dependency of LLM inference using LOOKAHEAD DECODING},
year = {2024},
publisher = {JMLR.org},
booktitle = {Proceedings of the 41st International Conference on Machine Learning},
articleno = {561},
numpages = {20},
location = {Vienna, Austria},
series = {ICML'24}
}

@inproceedings{
liu2025pearl,
title={{PEARL}: Parallel Speculative Decoding with Adaptive Draft Length},
author={Tianyu Liu and Yun Li and Qitan Lv and Kai Liu and Jianchen Zhu and Winston Hu and Xiao Sun},
booktitle={The Thirteenth International Conference on Learning Representations},
year={2025},
url={https://openreview.net/forum?id=QOXrVMiHGK}
}

@misc{li2025eaglespeculativesamplingrequires,
      title={EAGLE: Speculative Sampling Requires Rethinking Feature Uncertainty}, 
      author={Yuhui Li and Fangyun Wei and Chao Zhang and Hongyang Zhang},
      year={2025},
      eprint={2401.15077},
      archivePrefix={arXiv},
      primaryClass={cs.LG},
      url={https://arxiv.org/abs/2401.15077}, 
}

@inproceedings{li2024eagle2, 
	author = {Yuhui Li and Fangyun Wei and Chao Zhang and Hongyang Zhang}, 
	title = {{EAGLE-2}: Faster Inference of Language Models with Dynamic Draft Trees}, 
	booktitle = {Empirical Methods in Natural Language Processing},
	year = {2024}
}

@inproceedings{li2025eagle3,
    author = {Yuhui Li and Fangyun Wei and Chao Zhang and Hongyang Zhang},
    title = {{EAGLE-3}: Scaling up Inference Acceleration of Large Language Models via Training-Time Test}, 
    booktitle = {Annual Conference on Neural Information Processing Systems},
    year = {2025}
}

@misc{roziere2024codellamaopenfoundation,
      title={Code Llama: Open Foundation Models for Code}, 
      author={Baptiste Rozière and Jonas Gehring and Fabian Gloeckle and Sten Sootla and Itai Gat and Xiaoqing Ellen Tan and Yossi Adi and Jingyu Liu and Romain Sauvestre and Tal Remez and Jérémy Rapin and Artyom Kozhevnikov and Ivan Evtimov and Joanna Bitton and Manish Bhatt and Cristian Canton Ferrer and Aaron Grattafiori and Wenhan Xiong and Alexandre Défossez and Jade Copet and Faisal Azhar and Hugo Touvron and Louis Martin and Nicolas Usunier and Thomas Scialom and Gabriel Synnaeve},
      year={2024},
      eprint={2308.12950},
      archivePrefix={arXiv},
      primaryClass={cs.CL},
      url={https://arxiv.org/abs/2308.12950}, 
}

@misc{yang2025qwen3technicalreport,
      title={Qwen3 Technical Report}, 
      author={An Yang and Anfeng Li and Baosong Yang and Beichen Zhang and Binyuan Hui and Bo Zheng and Bowen Yu and Chang Gao and Chengen Huang and Chenxu Lv and Chujie Zheng and Dayiheng Liu and Fan Zhou and Fei Huang and Feng Hu and Hao Ge and Haoran Wei and Huan Lin and Jialong Tang and Jian Yang and Jianhong Tu and Jianwei Zhang and Jianxin Yang and Jiaxi Yang and Jing Zhou and Jingren Zhou and Junyang Lin and Kai Dang and Keqin Bao and Kexin Yang and Le Yu and Lianghao Deng and Mei Li and Mingfeng Xue and Mingze Li and Pei Zhang and Peng Wang and Qin Zhu and Rui Men and Ruize Gao and Shixuan Liu and Shuang Luo and Tianhao Li and Tianyi Tang and Wenbiao Yin and Xingzhang Ren and Xinyu Wang and Xinyu Zhang and Xuancheng Ren and Yang Fan and Yang Su and Yichang Zhang and Yinger Zhang and Yu Wan and Yuqiong Liu and Zekun Wang and Zeyu Cui and Zhenru Zhang and Zhipeng Zhou and Zihan Qiu},
      year={2025},
      eprint={2505.09388},
      archivePrefix={arXiv},
      primaryClass={cs.CL},
      url={https://arxiv.org/abs/2505.09388}, 
}

@dataset{sharegpt_vicuna_unfiltered,
  author = {Aeala},
  title = {ShareGPT\_Vicuna\_unfiltered},
  year = {2023},
  publisher = {Hugging Face},
  url = {https://huggingface.co/datasets/Aeala/ShareGPT_Vicuna_unfiltered}
}

@inproceedings{leviathan2023fast,
author = {Leviathan, Yaniv and Kalman, Matan and Matias, Yossi},
title = {Fast inference from transformers via speculative decoding},
year = {2023},
publisher = {JMLR.org},
booktitle = {Proceedings of the 40th International Conference on Machine Learning},
articleno = {795},
numpages = {13},
location = {Honolulu, Hawaii, USA},
series = {ICML'23}
}

@misc{chen2023acceleratinglargelanguagemodel,
      title={Accelerating Large Language Model Decoding with Speculative Sampling}, 
      author={Charlie Chen and Sebastian Borgeaud and Geoffrey Irving and Jean-Baptiste Lespiau and Laurent Sifre and John Jumper},
      year={2023},
      eprint={2302.01318},
      archivePrefix={arXiv},
      primaryClass={cs.CL},
      url={https://arxiv.org/abs/2302.01318}, 
}

@inproceedings{
lioubashevski2025looking,
title={Looking Beyond the Top-1: Transformers Determine Top Tokens in Order},
author={Daria Lioubashevski and Tomer M. Schlank and Gabriel Stanovsky and Ariel Goldstein},
booktitle={Forty-second International Conference on Machine Learning},
year={2025},
url={https://openreview.net/forum?id=2B11W1Z6ID}
}

@inproceedings{
csordas2026do,
title={Do Language Models Use Their Depth Efficiently?},
author={R{\'o}bert Csord{\'a}s and Christopher D Manning and Christopher Potts},
booktitle={The Thirty-ninth Annual Conference on Neural Information Processing Systems},
year={2026},
url={https://openreview.net/forum?id=Kz6eUL86XP}
}

@inproceedings{
Elbayad2020Depth-Adaptive,
title={Depth-Adaptive Transformer},
author={Maha Elbayad and Jiatao Gu and Edouard Grave and Michael Auli},
booktitle={International Conference on Learning Representations},
year={2020},
url={https://openreview.net/forum?id=SJg7KhVKPH}
}

@inproceedings{
schuster2022confident,
title={Confident Adaptive Language Modeling},
author={Tal Schuster and Adam Fisch and Jai Gupta and Mostafa Dehghani and Dara Bahri and Vinh Q. Tran and Yi Tay and Donald Metzler},
booktitle={Advances in Neural Information Processing Systems},
editor={Alice H. Oh and Alekh Agarwal and Danielle Belgrave and Kyunghyun Cho},
year={2022},
url={https://openreview.net/forum?id=uLYc4L3C81A}
}

@inproceedings{
chen2024eellm,
title={{EE}-{LLM}: Large-Scale Training and Inference of Early-Exit Large Language Models with 3D Parallelism},
author={Yanxi Chen and Xuchen Pan and Yaliang Li and Bolin Ding and Jingren Zhou},
booktitle={Forty-first International Conference on Machine Learning},
year={2024},
url={https://openreview.net/forum?id=xFk0w9zoV3}
}

@misc{raposo2024mixtureofdepthsdynamicallyallocatingcompute,
      title={Mixture-of-Depths: Dynamically allocating compute in transformer-based language models}, 
      author={David Raposo and Sam Ritter and Blake Richards and Timothy Lillicrap and Peter Conway Humphreys and Adam Santoro},
      year={2024},
      eprint={2404.02258},
      archivePrefix={arXiv},
      primaryClass={cs.LG},
      url={https://arxiv.org/abs/2404.02258}, 
}

@inproceedings{DBLP:conf/acl/Zhang00S0CM24,
  author       = {Jun Zhang and
                  Jue Wang and
                  Huan Li and
                  Lidan Shou and
                  Ke Chen and
                  Gang Chen and
                  Sharad Mehrotra},
  editor       = {Lun{-}Wei Ku and
                  Andre Martins and
                  Vivek Srikumar},
  title        = {Draft{\&} Verify: Lossless Large Language Model Acceleration via
                  Self-Speculative Decoding},
  booktitle    = {Proceedings of the 62nd Annual Meeting of the Association for Computational
                  Linguistics (Volume 1: Long Papers), {ACL} 2024, Bangkok, Thailand,
                  August 11-16, 2024},
  pages        = {11263--11282},
  publisher    = {Association for Computational Linguistics},
  year         = {2024},
  url          = {https://doi.org/10.18653/v1/2024.acl-long.607},
  doi          = {10.18653/V1/2024.ACL-LONG.607},
  bibsource    = {dblp computer science bibliography, https://dblp.org}
}

@misc{xia2025swiftontheflyselfspeculativedecoding,
      title={SWIFT: On-the-Fly Self-Speculative Decoding for LLM Inference Acceleration}, 
      author={Heming Xia and Yongqi Li and Jun Zhang and Cunxiao Du and Wenjie Li},
      year={2025},
      eprint={2410.06916},
      archivePrefix={arXiv},
      primaryClass={cs.CL},
      url={https://arxiv.org/abs/2410.06916}, 
}

@misc{
li2026optimized,
title={Optimized Early-Exit Based Speculative Decoding via Pipeline Parallelism},
author={Ruanjun Li and Ziheng Liu and Yuanming Shi and Jiawei Shao and Chi Zhang and Xuelong Li},
year={2026},
url={https://openreview.net/forum?id=6ezbdRe90k}
}

@misc{cai2024medusasimplellminference,
      title={Medusa: Simple LLM Inference Acceleration Framework with Multiple Decoding Heads}, 
      author={Tianle Cai and Yuhong Li and Zhengyang Geng and Hongwu Peng and Jason D. Lee and Deming Chen and Tri Dao},
      year={2024},
      eprint={2401.10774},
      archivePrefix={arXiv},
      primaryClass={cs.LG},
      url={https://arxiv.org/abs/2401.10774}, 
}

@misc{kumar2026speculativespeculativedecoding,
      title={Speculative Speculative Decoding}, 
      author={Tanishq Kumar and Tri Dao and Avner May},
      year={2026},
      eprint={2603.03251},
      archivePrefix={arXiv},
      primaryClass={cs.LG},
      url={https://arxiv.org/abs/2603.03251}, 
}

@misc{shoeybi2020megatronlmtrainingmultibillionparameter,
      title={Megatron-LM: Training Multi-Billion Parameter Language Models Using Model Parallelism}, 
      author={Mohammad Shoeybi and Mostofa Patwary and Raul Puri and Patrick LeGresley and Jared Casper and Bryan Catanzaro},
      year={2020},
      eprint={1909.08053},
      archivePrefix={arXiv},
      primaryClass={cs.CL},
      url={https://arxiv.org/abs/1909.08053}, 
}

@misc{huang2019gpipeefficienttraininggiant,
      title={GPipe: Efficient Training of Giant Neural Networks using Pipeline Parallelism}, 
      author={Yanping Huang and Youlong Cheng and Ankur Bapna and Orhan Firat and Mia Xu Chen and Dehao Chen and HyoukJoong Lee and Jiquan Ngiam and Quoc V. Le and Yonghui Wu and Zhifeng Chen},
      year={2019},
      eprint={1811.06965},
      archivePrefix={arXiv},
      primaryClass={cs.CV},
      url={https://arxiv.org/abs/1811.06965}, 
}

@misc{wu2025easyspeclayerparallelspeculativedecoding,
      title={EasySpec: Layer-Parallel Speculative Decoding for Efficient Multi-GPU Utilization}, 
      author={Yize Wu and Ke Gao and Ling Li and Yanjun Wu},
      year={2025},
      eprint={2502.02493},
      archivePrefix={arXiv},
      primaryClass={cs.LG},
      url={https://arxiv.org/abs/2502.02493}, 
}

%%%%%%%%%%%%%%%%%%%%%%%%%%%%%%%%%%%%%%%%%%%%%%%%%%%%%%%%%%%%

\appendix

\section{Additional formulation details and proofs}
\label{app:formulation}

Section~\ref{sec:prelim} presents the compact formulation used in the main paper. This appendix provides additional empirical measurements, illustrative examples, and full derivations for the selection and exploration results stated in the main text. We use the same notation, stable-EAD convention, and idealized layer-work accounting as in Section~\ref{sec:prelim}.

\subsection{Additional EAD measurements}
\label{app:ead-measurements}

\begin{figure}[t]
    \centering
    \includegraphics[width=\linewidth]{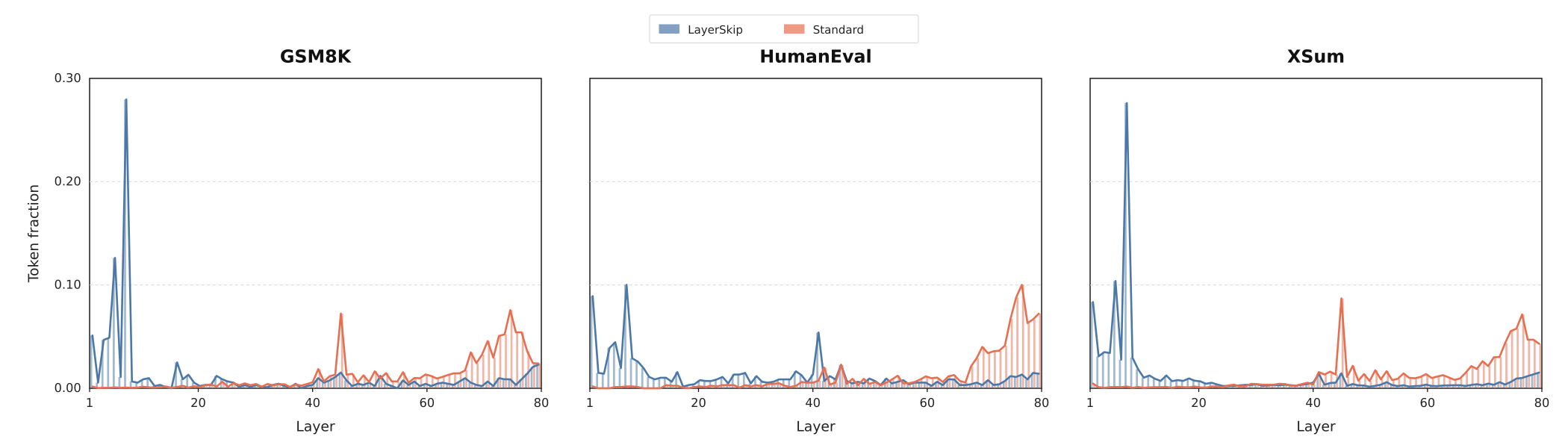}
    \caption{Distribution of stable EAD across different datasets on Llama2-70B.}
    \label{fig:ead}
\end{figure}

Using the stable EAD definition from Section~\ref{sec:prelim}, we measure token readiness on three datasets using two Llama2-70B variants: the standard model~\cite{touvron2023llama} and a LayerSkip model trained with early-exit supervision~\cite{elhoushi2024layerskip}. For visualization, we randomly sample 36 prompts per dataset and plot the EAD distributions in Figure~\ref{fig:ead}.

\noindent\textbf{Observation.}
The standard model has EADs concentrated near the final layer, while LayerSkip shows many tokens becoming stable at substantially earlier layers. This suggests that depth-side token readiness can be induced by early-exit-oriented training. The measurement motivates the adapter design in Section~\ref{sec:adapter}: adapters improve intermediate proposals by making non-final depths more likely to match the final-depth outcome, thereby shifting observed EADs earlier. The same measurement also motivates the decoding question addressed by DEX: given early-ready tokens, how should a lossless decoder exploit them without changing the final-depth decoding rule?

\subsection{Proof and example for selection overhead}
\label{app:selection-proof}

\paragraph{Proof of Proposition~\ref{prop:main-selection-overhead}.}
By the workload model in Section~\ref{sec:prelim}, the expected layer work charged to position $t$ under depth selection is
\begin{equation}
    W_t^{\mathrm{sel}}
    =p_t^{\mathrm{acc}}s_t
    +(1-p_t^{\mathrm{acc}})\left(L+\sum_{i=t}^{t+v_t-1}s_i\right).
\end{equation}
The EAD oracle would charge $\mathrm{EAD}(t)$ layers to the same token. Subtracting this oracle workload gives
\begin{align}
    c_t^{\mathrm{sel}}
    &=W_t^{\mathrm{sel}}-\mathrm{EAD}(t) \\
    &=p_t^{\mathrm{acc}}s_t
    +(1-p_t^{\mathrm{acc}})\left(L+\sum_{i=t}^{t+v_t-1}s_i\right)
    -\mathrm{EAD}(t) \\
    &=p_t^{\mathrm{acc}}(s_t-\mathrm{EAD}(t))
    +(1-p_t^{\mathrm{acc}})\left(L+\sum_{i=t}^{t+v_t-1}s_i-\mathrm{EAD}(t)\right),
\end{align}
where the last step uses $p_t^{\mathrm{acc}}+(1-p_t^{\mathrm{acc}})=1$. This is exactly the claimed decomposition. The first term is the overhead incurred when the selected exit succeeds, and the second term is the overhead incurred when the selected exit fails and descendant drafts are discarded.

\paragraph{Example.}
Consider an $L=20$ layer model and a token with $\mathrm{EAD}(t)=8$. A conservative selection $s_t=15$ reaches the token's stable-readiness depth, but it still pays $15-8=7$ extra layers beyond the EAD oracle. In contrast, an aggressive selection $s_t=4$ is earlier than the stable-readiness depth. If this exit fails and two descendant selected-exit tokens $s_{t+1}$ and $s_{t+2}$ are discarded, the failed branch charges
\begin{equation}
    20+4+s_{t+1}+s_{t+2}=24+s_{t+1}+s_{t+2}
\end{equation}
layer computations to position $t$ under the same accounting. This can exceed the $20$-layer cost of standard AR decoding for that position. The example illustrates why a single underestimated exit depth can turn an otherwise early-ready token into a full-depth fallback with additional rollback cost.

\subsection{Proof and example for exploration overhead}
\label{app:exploration-proof}

\paragraph{Proof of Proposition~\ref{prop:main-exploration-overhead}.}
By Definition~\ref{def:main-depth-exploration}, depth exploration charges token $t$ to the first explored depth that reaches the stable-readiness threshold:
\begin{equation}
    W_{\mathcal X}(t)=\left\lceil \mathrm{EAD}(t)\right\rceil_{\mathcal X}.
\end{equation}
Therefore
\begin{equation}
    c_t^{\mathrm{exp}}
    =W_{\mathcal X}(t)-\mathrm{EAD}(t)
    =\left\lceil \mathrm{EAD}(t)\right\rceil_{\mathcal X}-\mathrm{EAD}(t).
\end{equation}
The depth ceiling always returns an explored depth no earlier than its input, so $c_t^{\mathrm{exp}}\ge0$. By Definition~\ref{def:main-exploration-set}, $\Delta(\mathcal X)$ is the largest possible upward rounding error over all depths in $[L]$. Since $\mathrm{EAD}(t)\in[L]$, we have
\begin{equation}
    \left\lceil \mathrm{EAD}(t)\right\rceil_{\mathcal X}-\mathrm{EAD}(t)
    \le \Delta(\mathcal X).
\end{equation}
Thus $0\le c_t^{\mathrm{exp}}\le\Delta(\mathcal X)$.

\paragraph{Example.}
Using the same $L=20$ model and a token with $\mathrm{EAD}(t)=8$, consider the exploration set $\mathcal X=\{5,10,15,20\}$. Then
\begin{equation}
    \left\lceil \mathrm{EAD}(t)\right\rceil_{\mathcal X}
    =\left\lceil 8\right\rceil_{\mathcal X}=10.
\end{equation}
The earliest explored stable-ready branch is therefore the branch at depth $10$, which incurs only $10-8=2$ additional layers beyond the EAD oracle. The official output token remains the final-depth reference token; the selected explored branch only determines how much branch state can be preserved during lattice collapse. For this exploration set, the largest upward rounding error is $\Delta(\mathcal X)=4$, i.e., less than the adjacent spacing of $5$ layers. This deterministic rounding behavior contrasts with depth selection, where an underestimated exit depth can trigger full-depth fallback and rollback.

\subsection{Depth-side speedup derivation and monotonicity}
\label{app:speedup-proof}

\paragraph{Proof of Proposition~\ref{prop:main-speedup-monotonicity}.}
Standard AR decoding executes $L$ layers for each of $T$ generated tokens, so its total layer workload is $LT$. If a compared algorithm charges $W(t)$ layers to token $t$, its total charged layer workload is $\sum_{t=1}^{T}W(t)$. The corresponding depth-side speedup is therefore
\begin{equation}
    S=\frac{LT}{\sum_{t=1}^{T}W(t)},
\end{equation}
which gives Eq.~\eqref{eq:main-speedup-general}.

For depth exploration, substituting $W_{\mathcal X}(t)=\lceil\mathrm{EAD}(t)\rceil_{\mathcal X}$ gives
\begin{equation}
    S_{\mathcal X}
    =\frac{LT}{\sum_{t=1}^{T}W_{\mathcal X}(t)}
    =\frac{LT}{\sum_{t=1}^{T}\lceil\mathrm{EAD}(t)\rceil_{\mathcal X}},
\end{equation}
which gives Eq.~\eqref{eq:main-speedup-exploration}. Proposition~\ref{prop:main-exploration-overhead} implies
\begin{equation}
    \mathrm{EAD}(t)
    \le
    \left\lceil \mathrm{EAD}(t)\right\rceil_{\mathcal X}
    \le
    \min\{\mathrm{EAD}(t)+\Delta(\mathcal X),L\}.
\end{equation}
Summing over tokens gives
\begin{equation}
    \sum_{t=1}^{T}\mathrm{EAD}(t)
    \le
    \sum_{t=1}^{T}\left\lceil \mathrm{EAD}(t)\right\rceil_{\mathcal X}
    \le
    \sum_{t=1}^{T}\min\{\mathrm{EAD}(t)+\Delta(\mathcal X),L\}.
\end{equation}
All terms are positive, so taking reciprocals reverses the inequalities. Multiplying by $LT$ yields
\begin{equation}
    \frac{LT}{\sum_{t=1}^{T}\min\{\mathrm{EAD}(t)+\Delta(\mathcal X),L\}}
    \le S_{\mathcal X}
    \le
    \frac{LT}{\sum_{t=1}^{T}\mathrm{EAD}(t)}
    =S_{\mathrm{EAD}}.
\end{equation}
Since $\min\{\mathrm{EAD}(t)+\Delta(\mathcal X),L\}\le L$ for every $t$, the leftmost denominator is at most $LT$, so
\begin{equation}
    1\le
    \frac{LT}{\sum_{t=1}^{T}\min\{\mathrm{EAD}(t)+\Delta(\mathcal X),L\}}.
\end{equation}
Combining these inequalities gives Eq.~\eqref{eq:main-speedup-bound}.

It remains to prove monotonicity. If $\mathcal X\subseteq\mathcal X'$, then for every $\ell\in[L]$,
\begin{equation}
    \left\lceil \ell\right\rceil_{\mathcal X'}
    \le
    \left\lceil \ell\right\rceil_{\mathcal X},
\end{equation}
because the finer set $\mathcal X'$ has all depths available in $\mathcal X$ and possibly more. Therefore,
\begin{equation}
    \sum_{t=1}^{T}\left\lceil \mathrm{EAD}(t)\right\rceil_{\mathcal X'}
    \le
    \sum_{t=1}^{T}\left\lceil \mathrm{EAD}(t)\right\rceil_{\mathcal X}.
\end{equation}
Taking reciprocals and multiplying by $LT$ gives $S_{\mathcal X}\le S_{\mathcal X'}$. Since $\lceil\mathrm{EAD}(t)\rceil_{\mathcal X'}\ge\mathrm{EAD}(t)$ for every token, $S_{\mathcal X'}\le S_{\mathrm{EAD}}$. Hence $S_{\mathcal X}\le S_{\mathcal X'}\le S_{\mathrm{EAD}}$. Equality with the EAD oracle holds when $\mathcal X=[L]$.

\paragraph{Selection speedup for completeness.}
Substituting the selection workload from Proposition~\ref{prop:main-selection-overhead} into Eq.~\eqref{eq:main-speedup-general} gives
\begin{equation}
    S_{\mathrm{sel}}
    =
    \frac{LT}
    {\sum_{t=1}^{T}
    \left[
    p_t^{\mathrm{acc}}s_t
    +(1-p_t^{\mathrm{acc}})
    \left(
    L+\sum_{i=t}^{t+v_t-1}s_i
    \right)
    \right]}.
\end{equation}
This expression is not guaranteed to exceed one: failed selections increase the denominator through full-depth replacement and discarded descendant work. This is the speedup-side view of the selection bottleneck discussed in Section~\ref{sec:prelim}.

Together, these details support the main formulation. Depth selection compresses token-readiness uncertainty into a single selected exit depth, so its realized speedup depends on selector quality and rollback behavior. Depth exploration instead turns the unknown EAD into a deterministic upward rounding problem over $\mathcal X$, making the depth-side overhead explicitly controlled by exploration resolution. DEX instantiates this formulation by expanding multiple candidate depth branches, committing according to the final-depth reference, and preserving only branch states consistent with the committed prefix.

\section{\sys overhead analysis}
\label{app:overhead}

\begin{table}[t]
\centering
\small
\caption{
Overall runtime profile of DEX on the 8-explorer 70B run.
The table aggregates 160 DEX steps across 8 explorers/GPUs, giving 1280 rank-step samples.
Shares are normalized over the listed profiled components.
}
\label{tab:dex_all_rank_profile}
\begin{tabular}{lrr}
\toprule
Component & Avg. time / rank-step (ms) & Share of profiled time \\
\midrule
Expansion      & 13.830 & 89.13\% \\
Communication  &  1.222 &  7.88\% \\
Collapse       &  0.309 &  1.99\% \\
Decode/commit  &  0.156 &  1.01\% \\
\midrule
Total profiled & 15.517 & 100.00\% \\
\bottomrule
\end{tabular}
\end{table}

\noindent\textbf{From theoretical speed analysis to walltime speedup.}
Proposition~\ref{prop:main-speedup-monotonicity} characterizes the idealized depth-side critical-path opportunity of exploration. In the actual implementation, the final-depth token is retained as the exact commit reference, while parallel explorers compute reusable candidate branches before the commit is resolved. 
Thus, DEX accelerates decoding not by changing the target final-depth distribution, but by shortening the amount of post-commit computation that must be performed sequentially.

We use the following calibrated walltime model:
\[
S_{\mathcal X}^{\mathrm{wall}}
=
\frac{T(L+\delta_{\mathrm{AR}})}
{
\sum_{t=1}^{T}
\left(
\lceil \mathrm{EAD}(t)\rceil_{\mathcal X}
+
\delta_{\mathcal X}(t)
\right)
}.
\]
Here \(\delta_{\mathrm{AR}}\) denotes the average per-token overhead of AR beyond layer forward, and \(\delta_{\mathcal X}(t)\) denotes the extra cost paid by DEX at token \(t\) under exploration set \(\mathcal X\), including lattice expansion, batched branch execution, attention masking, KV-cache management, communication, commit, and collapse. Since DEX introduces branch and distributed execution costs that are not present in AR, the realized walltime speedup is lower than the idealized depth-side speedup. Our profile in Table~\ref{tab:dex_all_rank_profile} reports an implementation profile. Expansion remains the dominant cost, accounting for 89.13\% of the profiled time. The main non-expansion cost is communication, while collapse and decode/commit are small. This suggests that the practical runtime cost of DEX mainly comes from distributed synchronization and branch expansion, not from the commit--collapse bookkeeping itself.

\section{\sys additional details}
\label{app:additional_dex_method}
\subsection{DECS}
\label{app:decs}
Depth-Coupled Sampling (DECS) is used to diversify the intermediate
proposals exposed by different depths within the same sampling slot.
In the depth--position lattice, a sampling slot may be materialized at
multiple candidate depths, forming a slot chain. If each depth decodes
independently, shallower and deeper candidates in the same slot often
repeat the same token. Such duplicates provide little benefit: if the
repeated token matches the final-depth reference, the shallower occurrence
would already be selected; otherwise, repeating it only reduces the chance
that another depth proposes a token matching the reference.

For a lattice node \(i\), DECS collects the tokens that have already been
proposed by shallower nodes in the same slot chain, denoted by
\(\mathrm{Prior}(i)\), and masks these tokens before decoding the current
non-final candidate. This encourages deeper explorers to cover alternative
tokens while leaving the final-depth reference unchanged. In particular,
when the current node is the final-depth reference node
\(i_{\mathrm{ref}}\) at the final FSM state \(q_{\max}\), DECS disables
masking and decodes from the original logits. Therefore, the committed
token remains exactly the token produced by the unmodified final-depth
model, and DECS only affects which intermediate branch may be reused after
commitment.

For sampling, DECS uses a shared Gumbel buffer indexed by the slot root
\(r(i)\). All depths in the same slot chain use the same Gumbel row, so
the final-depth reference is sampled by the standard Gumbel-Max rule from
the unmasked final-depth logits. The masking operation is applied only to
non-final proposals and therefore does not change the target sampling
distribution of the committed token.

\begin{algorithm}[t]
\caption{Depth-Coupled Sampling (DECS)}
\label{alg:decs}
\begin{algorithmic}[1]
\Require Logits \(\ell\in\mathbb{R}^{V}\); lattice node \(i\); explore buffer \(\mathcal B\);
prior-depth map \(\mathrm{Prior}(\cdot)\); slot-root map \(r(\cdot)\);
mode \(m\in\{{greedy},{sample}\}\); temperature \(\tau\);
shared Gumbel buffer \(\mathcal G\); current state \(q\); final state \(q_{\max}\);
reference node \(i_{\mathrm{ref}}\)
\Ensure Decoded token \(y\)

\State \(\mathcal A \gets
\{\mathcal B[a]\mid a\in\mathrm{Prior}(i),\ \mathcal B[a]\neq\bot\}\)
\Comment{prior proposals in the same slot chain}

\If{\(q=q_{\max}\) and \(i=i_{\mathrm{ref}}\)}
    \State \(\mathcal A \gets \emptyset\)
    \Comment{preserve reference token}
\EndIf

\ForAll{\(v\in\mathcal A\)}
    \State \(\ell_v \gets -\infty\)
    \Comment{mask prior same-slot proposals}
\EndFor

\If{\(m={greedy}\)}
    \State \(y\gets \arg\max_v \ell_v\)
\Else
    \If{\(\mathcal G_{r(i),:}\) is uninitialized}
        \State Draw \(\mathcal G_{r(i),v}\sim\mathrm{Gumbel}(0,1)\) for all \(v\)
    \EndIf
    \State \(y\gets \arg\max_v\left(\ell_v/\tau+\mathcal G_{r(i),v}\right)\)
\EndIf

\State \(\mathcal B[i]\gets y\)
\State \Return \(y\)
\end{algorithmic}
\end{algorithm}

\subsection{Inducing adapter}
\label{app:adapter}

For standard LLMs, we train inducing adapters and keep the backbone fixed. 
Adapters are attached only to the middle candidate depths used by \sys, i.e., \(\ell\in\mathcal X\setminus\{L\}\); the final-depth branch uses the original hidden state and LM head without an adapter. 
In a DEX-\(1/K\) configuration, this corresponds to the \(K-1\) non-final stage boundaries. 
For a hidden state \(h_\ell\in\mathbb R^h\), the adapter computes
\[
\Delta h_\ell
=
\alpha_\ell W_{2,\ell}\sigma\!\left(W_{1,\ell}\mathrm{RMSNorm}(h_\ell)\right),
\qquad
\tilde h_\ell=h_\ell+\Delta h_\ell ,
\]
where \(W_{1,\ell}\in\mathbb R^{m\times h}\), \(W_{2,\ell}\in\mathbb R^{h\times m}\), \(m=2h\) by default, \(\sigma\) is the backbone activation, and \(\alpha_\ell\) is a learnable scalar initialized to \(10^{-3}\). 
The induced representation \(\tilde h_\ell\) is decoded by the same final normalization and LM head as the backbone.

The parameter count of one adapter is
\[
P_{\mathrm{ad}}
=
hm+mh+m+2h+1,
\]
including two linear layers with biases, the RMSNorm scale, and the scalar gate. 
With \(m=2h\), this becomes \(P_{\mathrm{ad}}=4h^2+4h+1\). 
For \(n=|\mathcal X|-1\) non-final depths, the total adapter size is \(nP_{\mathrm{ad}}\), and its relative size is \(nP_{\mathrm{ad}}/P_{\mathrm{backbone}}\). 
For example, this gives 536.9M parameters for CodeLlama-34B under DEX-\(1/3\) (\(\approx1.58\%\) of the backbone), 805.4M for Llama2-70B under DEX-\(1/4\) (\(\approx1.15\%\)), and 1.88B for Llama2-70B under DEX-\(1/8\) (\(\approx2.68\%\)).

We train the adapters by self-distillation on ShareGPT \cite{sharegpt_vicuna_unfiltered}, which contains around 70k samples for training.  We use 70K samples, truncate or pad each sequence to at most 1024 tokens, and mask padding and unsupervised positions from the loss. 
For each batch, the frozen full-depth model produces teacher logits \(z_L\). 
Each selected non-final depth produces student logits \(z_\ell\) through its adapter and the shared LM head. 
The objective is masked full-vocabulary KL distillation:
\[
\mathcal L
=
\frac{1}{|\mathcal X|-1}
\sum_{\ell\in\mathcal X\setminus\{L\}}
\tau^2
\mathrm{KL}
\left(
\mathrm{softmax}(z_L/\tau)
\;\|\;
\mathrm{softmax}(z_\ell/\tau)
\right),
\]
with \(\tau=1\). 

We optimize only adapter parameters with AdamW. 
The reported runs use learning rate \(10^{-3}\), weight decay 0, two epochs, 6\% warmup steps followed by linear decay, micro-batch size 4, and gradient accumulation 4, giving an effective batch size of 16. 
The backbone, final normalization, and LM head are frozen throughout training. 
Adapter training is performed on 4 NVIDIA H100 GPUs and takes about 8 hours for 32B/34B models and 12.6 hours for 70B models.

\section{Experiment details}
\label{app:eval}

\noindent\textbf{Implementations for each method.}
\sys is implemented with Hugging Face Transformers, and cross-GPU communication is performed through the NCCL backend.
Unless otherwise stated, all baselines are evaluated using their official implementations; Tensor Parallelism is launched with DeepSpeed.
For LayerSkip-trained models, \sys uses no inducing adapter.
For standard Llama/CodeLlama/Qwen models, \sys attaches inducing adapters only to non-final candidate depths, while the final-depth branch, final normalization, and LM head remain unchanged.
Thus, adapters only affect which intermediate branches may become reusable; the committed output is still resolved against the final-depth branch, preserving equivalence to the target decoding distribution.
PEARL uses Llama-2-7B as the draft model for Llama-2-70B, CodeLlama-7B for CodeLlama-34B, and Qwen3-1.7B for Qwen3-32B.
EAGLE-2 on Llama-2-70B uses \texttt{yuhuili/EAGLE-llama2-chat-70B} as the draft model, and EAGLE-3 on Qwen3-32B uses \texttt{AngelSlim/Qwen3-32B\_eagle3}.

\noindent\textbf{Hyperparameters.}
We tune LayerSkip self-speculation and AdaDecode on LayerSkip CodeLlama-34B and Llama-2-70B using five prompts from each of HumanEval, XSum, and GSM8K.
For self-speculation, the verification layer is fixed to the full model depth, i.e., 48 for CodeLlama-34B and 80 for Llama-2-70B.
For CodeLlama-34B, we search exit layers \(\{2,4,6,8,10\}\) and speculation lengths \(\{2,4,6,8\}\).
For Llama-2-70B, we search exit layers \(\{4,8,12,16,20\}\) with the same speculation lengths.
For AdaDecode, intermediate hidden states are projected using the shared LayerSkip LM head, with shortcut layers uniformly distributed across depth: layers \(8,16,24,32,40\) for CodeLlama-34B and layers \(10,20,\ldots,70\) for Llama-2-70B.
We search speculation lengths \(\{2,4,5,6,8\}\) and confidence thresholds \(\{0.2,0.4,0.6,0.8\}\).
Throughput is measured as the total number of generated tokens divided by the total end-to-end response time.

The final self-speculation settings are as follows.
For CodeLlama-34B, we use HumanEval \((8,4)\), GSM8K \((2,2)\), and XSum \((10,4)\).
For Llama-2-70B, we use HumanEval \((8,4)\), GSM8K \((8,4)\), and XSum \((8,6)\).
Each tuple denotes \((\texttt{exit\_layer}, \texttt{num\_speculations})\).

The final AdaDecode settings are as follows.
For CodeLlama-34B, we use HumanEval \((8,0.6)\), GSM8K \((8,0.6)\), and XSum \((8,0.8)\).
For Llama-2-70B, we use HumanEval \((8,0.8)\), GSM8K \((8,0.8)\), and XSum \((8,0.6)\).
Each tuple denotes \((\texttt{num\_speculations}, \texttt{threshold})\).

For DEL, we use its official dynamic decoding configuration and do not perform an additional grid search.

For Lookahead, PEARL, and EAGLE-series baselines, we follow their official evaluation configurations.

\noindent\textbf{Datasets.}
We randomly sample 128 prompts from each of GSM8K, HumanEval, and XSum.
For Llama and CodeLlama experiments, the maximum number of new tokens is set to 512.
For Qwen models, we set the maximum number of new tokens to 1024 to avoid truncating longer generated responses.
By default, evaluations use greedy decoding with batch size 1.

\section{Limitations and discussion}
\label{app:discussion}

\noindent\textbf{Hardware-agnostic formulation and resource-dependent realization.}
Depth exploration separates the algorithmic question of \emph{which depths to expose} from the systems question of \emph{how to execute them}. 
The exploration set \(\mathcal X\), the rounded workload \(W_{\mathcal X}(t)=\lceil \mathrm{EAD}(t)\rceil_{\mathcal X}\), and the resolution bound in Proposition~\ref{prop:main-speedup-monotonicity} characterize, under idealized layer-work accounting, how a finite set of candidate depths approximates the EAD oracle. 
This formulation does not prescribe a hardware schedule: the same lattice could be evaluated sequentially, or with limited opportunistic overlap, on a single device, but such schedules mainly expose additional candidate states and are not expected to provide wall-clock speedup without sufficient parallel execution resources. 

\sys is one distributed realization of this formulation. 
To convert depth exploration into wall-clock speedup, our current implementation assigns each explored depth boundary to a depth explorer and places each explorer on a separate GPU. 
Thus, a \sys-1/K configuration uses \(K\) depth explorers and \(K\) GPUs. 
Additional GPUs instantiate finer exploration resolution rather than serving as hidden resources. 
We therefore report GPU counts explicitly and separate matched-resource comparisons from scaling results. 
For resource-aware comparison, we evaluate tensor parallelism with 2, 4, and 8 GPUs and PEARL under its distributed configuration, using 3 GPUs for 32B/34B models and 4 GPUs for the 70B model. 
Under matched GPU counts, \sys is competitive with PEARL on 34B and 70B models, while higher-\(K\) configurations evaluate how additional devices translate into finer exploration resolution and higher wall-clock throughput. 

\noindent\textbf{Expansion growth and overhead.}
As described in Section~\ref{sec:lattice_fsm}, \sys expands a depth--position lattice during each exploration cycle. With \(K\) depth explorers, a full expansion cycle can materialize up to \(2^K-1\) exploration tokens, and the largest token batch processed by one explorer in our 8-explorer configuration reaches \(2^7=128\) tokens. This growth is the main practical cost of finer exploration. While Proposition~\ref{prop:main-speedup-monotonicity} shows that finer exploration reduces the idealized depth-side rounding error, the realized wall-clock gain also depends on explorer forward cost, memory traffic, attention-mask construction, communication, and branch-state management. Our current scaling study evaluates up to 8 explorers, where \sys continues to improve in the tested settings. We do not extrapolate this monotonic wall-clock behavior beyond the evaluated range: at larger \(K\), marginal gains may saturate as lattice expansion and compute-bound overheads become more prominent. A promising direction is to prune low-value branches or construct sparse exploration lattices while preserving the final-depth commit invariant.

\noindent\textbf{Resource-dependent limit of pure depth exploration.}
Pure depth exploration also has a resource-dependent upper bound. Let \(d_0\) be the shallowest explored boundary in \(\mathcal X\). Even if every token becomes ready before \(d_0\), \sys can only reuse the branch at \(d_0\), so the idealized depth-side speedup is bounded by approximately \(L/d_0\). Under a uniform \(K\)-way depth partition, \(d_0 \approx L/K\), giving an approximate upper bound of \(K\), which matches the number of depth explorers in the current implementation. This limit clarifies the role of exploration resolution: increasing \(K\) improves the depth-side approximation to the EAD oracle, but also requires more devices and increases lattice-expansion overhead. Extending \sys beyond pure depth exploration, for example by combining depth exploration with token-axis speculative decoding, is an orthogonal direction. Such a design would require coordinating token-tree verification with depth-lattice collapse and is left to future work.

\paragraph{Broader impacts.}
DEX improves the efficiency of autoregressive LLM decoding by increasing end-to-end throughput while preserving the final-depth decoding rule. This can reduce inference latency per generated token, making large-model deployment more accessible. At the same time, faster and cheaper generation may also lower the cost of harmful large-scale uses, such as spam, misinformation, or automated abuse. Because DEX is a lossless decoding wrapper and does not modify the final-depth model distribution, these risks are primarily inherited from the underlying model and deployment context. Practical deployments should therefore combine decoding acceleration with the same safety filtering, monitoring, and rate-limiting mechanisms used for the base model.
% \section{Technical appendices and supplementary material}
% Technical appendices with additional results, figures, graphs, and proofs may be submitted with the paper submission before the full submission deadline (see above). You can upload a ZIP file for videos or code, but do not upload a separate PDF file for the appendix. There is no page limit for the technical appendices. 

% Note: Think of the appendix as ``optional reading'' for reviewers. The paper must be able to stand alone without the appendix; for example, adding critical experiments that support the main claims to an appendix is inappropriate. 

%%%%%%%%%%%%%%%%%%%%%%%%%%%%%%%%%%%%%%%%%%%%%%%%%%%%%%%%%%%%

% \newpage
% \input{sections/7-checklist}

\end{document}